%% file: main.tex
\documentclass[11pt,letterpaper]{article}
\usepackage{arxiv_preprint}
\usepackage{amsmath,amssymb,amsfonts}
\usepackage{graphicx,booktabs,tabularx,array}
\usepackage{microtype}
\usepackage{placeins}
\usepackage{float}
\floatstyle{ruled}
\newfloat{algorithm}{tbp}{loa}
\floatname{algorithm}{Algorithm}
\usepackage[hidelinks,bookmarksnumbered]{hyperref}
\usepackage{url}
\newcommand{\agent}{PACE}
\newcommand{\bench}{ResearchProofBench}
\newcolumntype{Y}{>{\raggedright\arraybackslash}X}
\newtheorem{theorem}{Theorem}

\title{\agent: Research Proving Agent for Adaptive Conjecture Exploration}
\input{authors.tex}
\date{}

\hypersetup{pdftitle={PACE: Research Proving Agent for Adaptive Conjecture Exploration},pdfauthor={Shu Miao, Feiming Wang, Wentao Long, Chenyi Li, Kun Yuan, Zaiwen Wen},pdfsubject={Research preprint}}

\begin{document}
\maketitle
\begin{abstract}
Many mathematical research agents focus on solving problems with fixed
statements. However, conjecture research, especially in applied mathematics,
may require revising candidate questions or conclusions as investigation
proceeds, coupling problem formulation with proof development. We present
\agent, a research--proof agent that supports this process through a
structured research loop that selects mathematical objectives and develops
or challenges candidate mechanisms. The whole prover constructs proofs from
these mechanisms, and an independent verifier checks them against the
original target. Evidence-tracking memory stores assumptions and dependencies
arising during research and marks evidence for revalidation when a candidate
or condition changes. This memory supports task-specific context composition
that keeps checked premises, exploratory material and unresolved obligations
distinct. To evaluate proof construction and conjecture research, we introduce
\bench, an applied-mathematics benchmark focused on optimization, comprising
51 known-proof tasks with 331 hidden checkpoints under historical abstract
access and six open-problem cases. The known-proof tasks assess reconstruction
of published results, while the open-problem cases assess mathematical
investigation and proof development. On the 51 known-proof tasks,
\agent\ achieves an aggregate Proof Score of 75.16 and completes 20 proofs.
In conjecture research, \agent\ develops conditional results, with its
submissions receiving higher review scores than the compared baselines.
\end{abstract}

\section{Introduction}
\label{sec:introduction}

Generating natural-language proofs for research-level mathematics requires
coordinating mathematical reasoning, relevant literature and statement revision. NaturalProofs studies retrieval of mathematical references, while
NaturalProver uses retrieved or supplied definitions and theorems to guide
stepwise and complete proof generation
\citep{welleck2021naturalproofs,welleck2022naturalprover}. ProofBench and
ProofGrader study graded evaluation and selection of generated proofs using
expert annotations and problem-specific marking schemes
\citep{ma2026proofbench}. Human--AI workflows combine expert guidance on
problem formulation with model-assisted exploration of conjectures, proofs
and counterexamples
\citep{davies2021intuition,charton2024patternboost,georgiev2025exploration,li2025interactive,lai2025grover}.
Computational approaches also generate candidate conjectures
\citep{raayoni2021ramanujan,davila2024txgraffiti} and search for mathematical
constructions and counterexamples
\citep{wagner2021constructions,romeraparedes2024funsearch,novikov2025alphaevolve}.
AI co-mathematician provides an
interactive workspace for literature search, computational exploration and
theorem proving, retaining project state and failed hypotheses across
investigations \citep{zheng2026aicomath}. These activities require repeated
decisions about what to investigate, which results to reuse and how to revise
an argument. Mathematical research agents coordinate these decisions within
workflows for proof development:
QED separates planning, proving and verification for open problems
\citep{an2026qed}. Aletheia combines iterative proof generation, verification
and revision with literature search \citep{feng2026aletheia}. Iteris connects
proof development with numerical exploration and mathematical constructions
\citep{chen2026iteris}. Related work on agents studies the action selection, feedback and skill
reuse underlying such workflows \citep{wang2023voyager}.
ReAct interleaves reasoning with actions \citep{yao2023react},
and AutoGen supports collaboration through configurable conversations between
agents with access to tools and human input \citep{wu2023autogen}.
Self-Refine alternates feedback and revision within a single model
\citep{madaan2023selfrefine}, while Reflexion retains reflections to guide
subsequent attempts \citep{shinn2023reflexion}. For longer investigations,
Agent Workflow Memory extracts reusable procedures
\citep{wang2024workflow}, Context-Folding summarizes completed subtask
trajectories to limit the active context \citep{sun2025contextfolding},
and agentic context engineering maintains structured context through
generation, reflection and curation \citep{zhang2025ace}.

\begin{figure}[t]
  \centering
  \includegraphics[width=\linewidth]{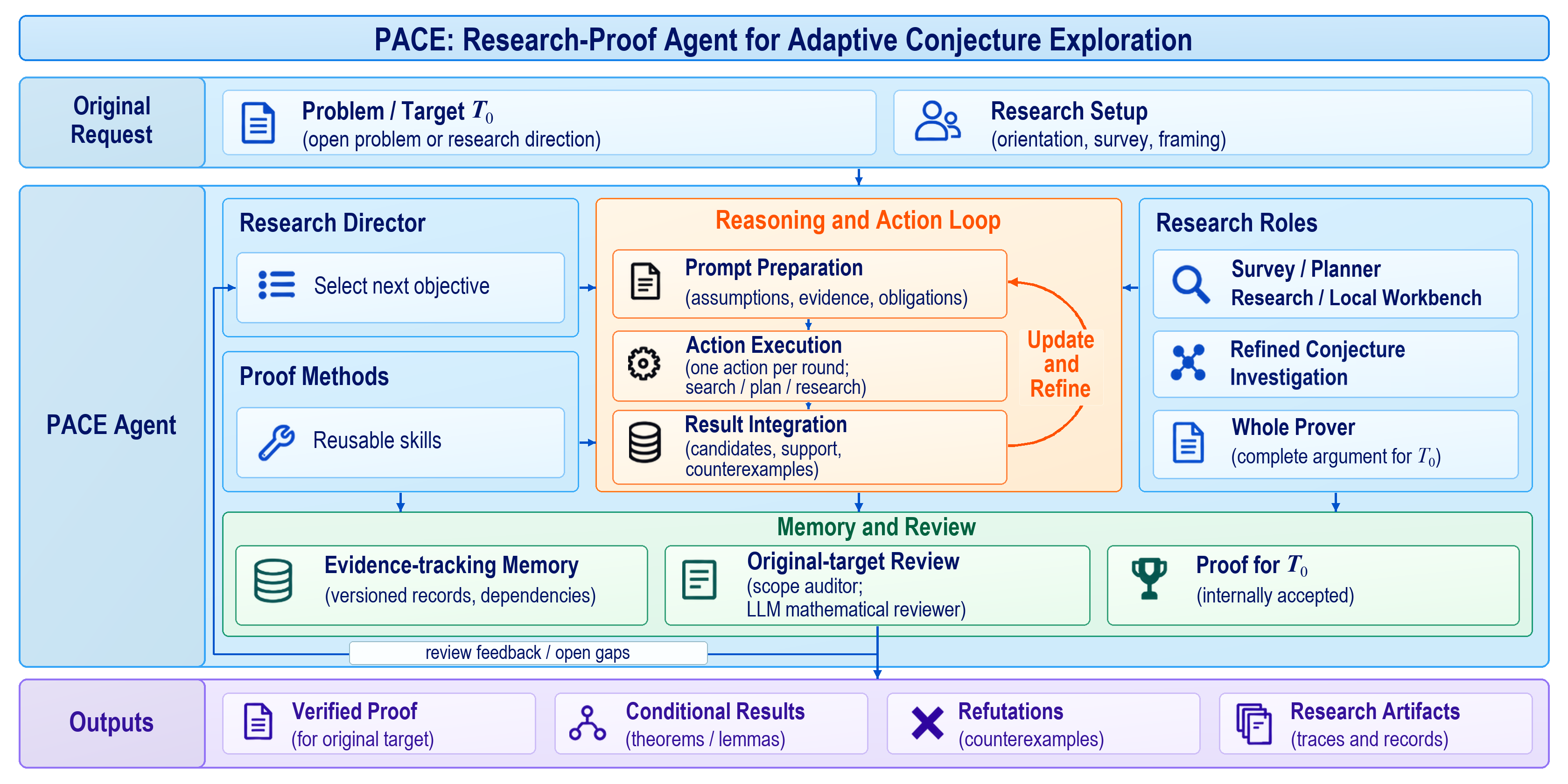}
  \caption{Overview of \agent\ for direct proving and conjecture research.}
  \label{fig:overview}
\end{figure}

Conjecture research requires exploring which statements can be established
and under what conditions. Partial derivations and counterexamples can
reveal subproblems, motivate candidate refinement or redirect proof
strategies. As objectives evolve, a lemma may remain valid without
supporting the revised candidate. A conditional theorem may leave part
of the original question unresolved. We study how to develop candidates
and proofs together, using local findings to guide exploration while
reassessing support affected by revision. We present \agent, which organizes this exploration through refined
conjecture investigation (RCI) within a research program
(Figure~\ref{fig:overview}). RCI develops research seeds into falsifiable
candidates and supporting mechanisms. Proof and refutation probes guide
candidate refinement, splitting and mechanism repair. Local results and
unresolved obligations inform the next objective. Evidence-tracking memory
retains assumptions, dependencies and version-specific assessments, and the
prompt composer separates applicable premises from exploratory material
and remaining obligations, with affected support revalidated before reuse.
Local reasoning and the whole prover develop candidate proofs for
independent review
(Table~\ref{tab:mechanism-comparison}).

To evaluate proof construction and conjecture exploration, we introduce
\bench, an applied-mathematics benchmark focused on optimization, with
51 known-proof tasks from 24 papers and six open-problem cases. Known-proof
evaluates agents' ability to reconstruct solutions from task statements and
search results containing historical literature abstracts, using 331
hidden checkpoints to assess proof completeness. The open-problem cases
assess conditional theorems, bounds and counterexamples against the original
objectives. Independent review measures completion of the registered mathematical
obligations in known-proof and progress toward the original objectives in the
open-problem cases.
Known-proof search ablations compare no-web, frozen and full-web access. Our
contributions are summarized as follows.

\begin{table}[!t]
\centering\small\setlength{\tabcolsep}{4pt}
\renewcommand{\arraystretch}{1.15}
\caption{Research mechanisms in QED~\citep{an2026qed},
Iteris~\citep{chen2026iteris} and \agent.}
\label{tab:mechanism-comparison}
\begin{tabularx}{\linewidth}{@{}l*{5}{>{\centering\arraybackslash}X}@{}}
\toprule
System & \shortstack{Plan\\subproblems}
& \shortstack{Refine\\conjectures}
& \shortstack{Record result\\conditions}
& \shortstack{Recheck support\\after revisions}
& \shortstack{Separate prompt\\composition}\\
\midrule
QED & $\checkmark$ & --- & $\checkmark$ & --- & ---\\
Iteris & $\checkmark$ & $\checkmark$ & $\checkmark$ & --- & ---\\
\textbf{\agent} & $\checkmark$ & $\checkmark$ & $\checkmark$ & $\checkmark$ & $\checkmark$\\
\bottomrule
\end{tabularx}
\end{table}

\textbf{(i) Evidence-tracking memory and task-specific prompt composition.}
We organize research claims and supporting evidence as linked records with explicit scope, dependencies and review status. A task-specific prompt composer assembles these records into mathematical context that distinguishes checked premises, exploratory material and unresolved obligations, preserving the conditions under which prior results can support the current objective.

\textbf{(ii) Research programs and refined conjecture investigation.}
We organize conjecture research as a research program of mathematical subproblems with explicit objectives and dependencies. Within each item, refined conjecture investigation (RCI) develops research seeds into candidate tasks, refines their scope and constructs supporting mechanisms. Local results and unresolved bridges then guide subsequent planning and derivations.

\textbf{(iii) Condition-sensitive evaluation of proofs and conjecture research.}
We introduce \bench\ for research-level proof workflows in applied
mathematics, with an emphasis on optimization. Its 51 statement-only
known-proof tasks use 331 hidden checkpoints and controlled access to
historical literature abstracts. Independent review assesses
graded progress, fully completed obligations and complete proofs under the task rubric,
accounting for assumptions and dependencies. Six optimization-related
conjecture cases extend
evaluation to conditional results, bounds and counterexamples against the
original research objectives.

\section{Problem Formulation and Preliminaries}
\label{sec:preliminaries}

\subsection{Mathematical Task Formulation}
Let $T_0$ denote the original research request, including its mathematical
scope and completion criteria. The agent supports two modes. In direct
proof, the proposition is fixed while intermediate claims and proof
mechanisms may change. In research program, candidate statements and their
conditions may also evolve within the alternatives allowed by $T_0$.
Both modes retain the original completion criteria. At step $t$, the
mathematical state is
\begin{equation}
s_t=(T_0,W_t,M_t,\mathcal R_t,\mathcal O_t),
\label{eq:state}
\end{equation}
where $W_t$ is the dependency graph of mathematical subproblems, $M_t$ is
evidence-tracking memory, $\mathcal R_t$ contains current candidate and
mechanism versions with their assessments, and $\mathcal O_t$ contains
unresolved derivations, applicability checks and coverage obligations.
A revision updates the relevant versions and reopens affected support
checks while retaining earlier results in $M_t$ under their established
conditions.

\subsection{Research outcomes and completion}
\label{sec:research-outcomes}
A universally quantified candidate $q$ has the form
$\forall u\in\mathcal X_q,\ \Gamma_q(u)\Rightarrow\varphi_q(u)$,
with assumptions $\Gamma_q$ and conclusion $\varphi_q$.
Its supporting claims $\mathcal H_q=\{\psi_1,\ldots,\psi_m\}$ retain their
hypotheses, dependencies and coverage of $T_0$. A refutation provides a
counterexample $u^\star\in\mathcal X_q$ satisfying
$\Gamma_q(u^\star)\land\neg\varphi_q(u^\star)$. Construction tasks preserve
their quantifier order (Appendix~\ref{app:task-forms}). Missing derivations,
applicability bridges and uncovered target clauses remain in $\mathcal O_t$.
Applying a source theorem $H_s(v)\Rightarrow C_s(v)$ to a goal $\psi(u)$
under assumptions $\Delta$ requires an instantiation $v=\theta(u)$ satisfying
$\Delta\vdash H_s(\theta(u))$ and
$\Delta\cup\{C_s(\theta(u))\}\vdash\psi(u)$.
Let $\mathcal E_q$ denote applicable source results and order $\mathcal H_q$
by dependency. Review checks support through
$\Gamma_q\cup\mathcal E_q\cup\{\psi_j:j<i\}\vdash\psi_i$ for $i=1,\ldots,m$,
and candidate coverage through
$\Gamma_q\cup\mathcal E_q\cup\mathcal H_q\vdash\varphi_q$. These checks
refer to the current candidate and mechanism: a changed hypothesis can
invalidate a source application without invalidating the source theorem.

Completion also requires relating the proved candidate to the original
request. For a final argument $\pi$, the auditor $A$ checks the deliverables
required by $T_0$, and the mathematical reviewer $V$ checks derivations,
source applications and compatibility of assumptions. Internal acceptance is
$\operatorname{Acc}(T_0,\pi)=1\Longleftrightarrow A(T_0,\pi)=1\land V(T_0,\pi)=1$.
A conditional result completes an obligation only if $T_0$ permits that
endpoint or its additional conditions are established. Otherwise it remains
partial progress. Proving a narrower candidate leaves the uncovered part
of $T_0$ unresolved.

\section{Methodology for conjecture research}
\label{sec:method}
\agent\ integrates evidence-tracking memory, task-specific prompt composition
and refined conjecture investigation into a research loop that coordinates
evidence reuse, candidate refinement and proof development
(Figure~\ref{fig:overview}).

\subsection{Evidence-tracking memory for task-specific prompt composition}
\label{sec:seg-memory}
\textbf{Results and their support relations.}
Evidence-tracking memory $M_t$ separates results from their uses in a graph $\mathcal G_t=(V_t,E_t)$. Each node is a versioned research record,
such as a candidate claim, supporting derivation or unresolved obligation,
with scope described by
$\sigma(v)=(\mathcal C_v,\Gamma_v,Q_v,\Omega_v)$. $\mathcal C_v$ specifies the mathematical model,
$\Gamma_v$ the hypotheses, $Q_v$ the quantifier pattern and $\Omega_v$ the
covered region and output. Edges distinguish three relations:
$\mathrm{supports}$ connects evidence to a candidate,
$\mathrm{depends\_on}$ connects a result to its premises and
$\mathrm{reopens}$ identifies an assessment requiring renewed verification
after a revision. Support assessments refer to the current candidate and
mechanism versions. The results retain their own conditions and review findings,
so invalidating a use does not mean to deny the underlying result for future reuse.

\textbf{Applicability after revision.}
To reuse a proved conditional result $\Gamma_v\Rightarrow\psi_v$ for a
revised candidate $q'$, instantiate it in the candidate's setting and check
$\Gamma_{q'}\Rightarrow\Gamma_v$, together with the derivation connecting
$\psi_v$ to its intended use. Quantifiers, outputs and the covered region
must match that application. A revision creates a new record and reopens
potentially affected support assessments identified through the recorded
dependencies. Reopened assessments require renewed mathematical review
before their conclusions can serve as established support for the revised
argument. Historical records retain their original scope and review
findings. Unaffected results remain available under their established
conditions. Missing applicability bridges enter $\mathcal O_t$. Recorded
dependencies identify checks to reopen and mathematical reasoning resolves them.

\textbf{Context construction for a proof objective.}
For the next objective $g$, the prompt composer organizes selected checked premises $\mathcal P_t(g)$, exploratory material
$\mathcal U_t(g)$ and relevant unresolved obligations $\mathcal O_t(g)$,
forming $\operatorname{Ctx}_t(g)=\langle g,\mathcal P_t(g),\mathcal U_t(g),\mathcal O_t(g)\rangle$.
A checked conditional lemma enters with its assumptions and its conclusion
can be used only after assumptions are established in the current
setting. An applicable conclusion remains
available for exploration, with the missing justification in
$\mathcal O_t(g)$. It cannot enter $\mathcal P_t(g)$ as an established
premise for that use. Failed routes can therefore inform investigation
without inheriting their former support status. Method guidance supplies
reasoning strategies. The composition policy appears in
Appendix~\ref{app:composer}.

\subsection{Research programs with refined conjecture investigation}
\label{sec:loop}
\label{sec:smi}
\label{sec:hierarchy}
\textbf{Research objectives from unresolved obligations.}
A research program $W_t$ decomposes the original request $T_0$ into
mathematical objectives within its scope, specifying their dependencies and
contributions to its completion criteria and the original request $T_0$. An unresolved derivation, an
applicability condition or a gap in target coverage can each become a
subproblem. The next objective is selected from $\mathcal O_t$ using the
currently applicable support. Revision redirects research by exposing
bridges needed to reuse established results. Combining those results
requires compatible assumptions and coverage of the original deliverables.

\textbf{Coupled candidate and mechanism refinement.}
Within a subproblem, RCI develops a pair $(q,\mathcal H_q)$ comprising a
candidate statement and its proposed supporting claims
(Figure~\ref{fig:smi}). The candidate specifies the assumptions, conclusion
and refutation target. The mechanism proposes intermediate claims sufficient
to establish it. Seed assessment identifies the mathematical content beyond
available sources. Support and coverage checks then distinguish whether
the intermediate claims are justified without circular use of $q$ and
whether they establish the entire candidate. Proof and refutation probes
identify the object requiring change: a failed intermediate argument can
leave $q$ open, whereas a counterexample satisfying its assumptions challenges
the candidate itself. RCI uses this distinction to guide three transitions:
\begingroup
\setlength{\abovedisplayskip}{6pt}
\setlength{\belowdisplayskip}{6pt}
\setlength{\abovedisplayshortskip}{6pt}
\setlength{\belowdisplayshortskip}{6pt}
\[
(q,\mathcal H_q)\longrightarrow
\begin{cases}
(q,\mathcal H'_q), & \text{repair the supporting argument},\\
(q',\mathcal H_{q'}), & \text{revise the candidate statement},\\
\{(q_i,\mathcal H_{q_i})\}_{i=1}^{2}, & \text{split the candidate into subproblems}.
\end{cases}
\]
\endgroup
These transitions reopen different obligations. A mechanism repair preserves
$q$ while rechecking changed support and affected coverage. Candidate
revision requires reassessing earlier support and counterexamples under
the new assumptions, conclusion and domain. Splitting requires justifying
the subproblems and their combined coverage of the parent objective.
Earlier results retain their original conditions and affected support must be
revalidated under Section~\ref{sec:seg-memory}. Appendix~\ref{app:l0l1-running-example}
illustrates this reuse check with a local curvature lemma
(Figure~\ref{fig:revision-support}). Revisions stay within the alternatives allowed by $T_0$ and
a restricted conclusion leaves remaining coverage explicit in $\mathcal O_t$.

\begingroup
\setlength{\intextsep}{10pt}
\begin{figure}[htbp]
  \centering
  \includegraphics[width=\linewidth]{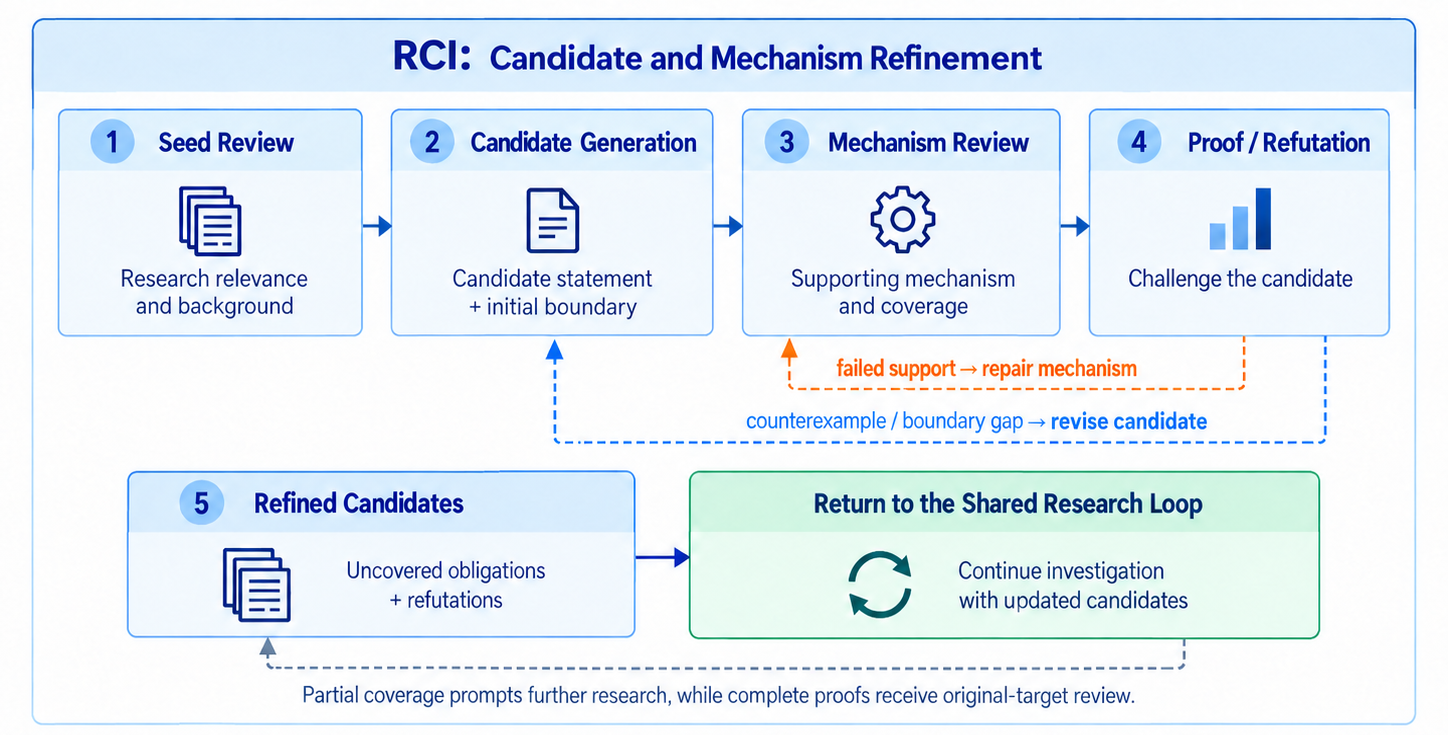}
  \caption{Candidate and mechanism refinement in RCI.}
  \label{fig:smi}
\end{figure}
\endgroup

\textbf{From revised support to proof development.}
RCI returns $(q,\mathcal H_q,\mathcal O_q)$, retaining local support together
with the obligations needed to use it. Unestablished assumptions of a
conditional result become further proof objectives. The local workbench
addresses these derivations or tests boundary cases, and the whole prover
combines applicable results into a candidate proof. Joint review checks
derivations and coverage against $T_0$ under
Section~\ref{sec:research-outcomes}. Both assessments concern the same
current argument, so revision requires renewed acceptance. Remaining gaps
return to $\mathcal O_t$ for further derivation or candidate refinement.
Detailed RCI phases appear in Appendix~\ref{app:smi}.

Algorithm~\ref{alg:research-program} summarizes this research loop over at
most $\tau$ investigation steps. At step $t$, the agent selects an objective
$g_t$ and an action $a_t$, using $\operatorname{Ctx}_t(g_t)$ to obtain a result
$z_t$. The result and any proof-review findings identify discharged
obligations $\mathcal D_t$, new obligations $\mathcal N_t$ and obligations
$\mathcal B_t$ reopened by revision. These findings are recorded in the next
state $s_{t+1}$ before composing another context. Reopened obligations can
therefore guide further research while earlier arguments remain available
in memory.

\begin{algorithm}[htbp]
\caption{Research program and proof development}
\label{alg:research-program}
\begin{tabularx}{\linewidth}{@{}rX@{}}
\multicolumn{2}{@{}l}{\textbf{Input:} original target $T_0$, initial memory $M_0$, investigation limit $\tau$.}\\
1. & Frame $T_0$ into subproblems $W_0$ and initialize $s_0$ with $M_0$.\\
2. & \textbf{for} $t=0,\ldots,\tau-1$ \textbf{do}\\
3. & \quad Select objective $g_t$ and action $a_t$ from $s_t$;
     execute $a_t$ using $\operatorname{Ctx}_t(g_t)$ to obtain $z_t$.\\
4. & \quad If $z_t$ is a complete candidate proof, have the auditor and
     mathematical reviewer assess it against $T_0$ and record their findings.\\
5. & \quad Identify $\mathcal D_t$, $\mathcal N_t$ and $\mathcal B_t$
     from the result, current support assessments and any review findings.\\
6. & \quad $\mathcal O_{t+1}\leftarrow
     (\mathcal O_t\setminus\mathcal D_t)\cup\mathcal N_t\cup\mathcal B_t$.\\
7. & \quad Update the research program, memory, and candidate and mechanism records;
     form $s_{t+1}$ with unresolved obligations $\mathcal O_{t+1}$.\\
8. & \quad \textbf{if} $z_t$ is a complete candidate proof and
     $\operatorname{Acc}(T_0,z_t)=1$ \textbf{then return} $z_t$.\\
9. & \textbf{end for}\\
10. & \textbf{return} $(M_\tau,\mathcal O_\tau)$.\\
\end{tabularx}
\end{algorithm}
\FloatBarrier

\subsection{Example: composite optimization under generalized smoothness}
\label{sec:conjecture-example}
\textbf{1. Candidate formulation.}
Consider $F=f+h$ under $(L_0,L_1)$ smoothness, with convex $f$ and convex
proximable $h$ whose subgradients are bounded by $\ell$. Under the
strong, global subgradient assumption, RCI seeks an algorithm with accelerated leading order
$R\sqrt{L_{\mathrm{eff}}/\varepsilon}$ uniformly over instances, where
$L_{\mathrm{eff}}=L_0+L_1\ell$, $R$ bounds the initial distance to a
minimizer and $\varepsilon$ is the target accuracy. The task permits a
precise conditional conclusion. Full records appear in
Appendix~\ref{app:l0l1-running-example}.

\textbf{2. Retaining a result after mechanism failure.}
The initial mechanism treats curvature as bounded by an absolute multiple
of $L_{\mathrm{eff}}$. A refutation probe constructs an admissible
one-dimensional example with exponentially growing curvature, invalidating
that step while leaving the candidate open. A local bound remains available:
$\bar L_\rho=e^{L_1\rho}L_{\mathrm{eff}}$ within distance $\rho$ of a
minimizer. Its mathematical validity does not establish its applicability
along the algorithm's trajectory. The query points, proximal outputs and
relevant segments must stay in that region, and the desired complexity
requires $L_1\rho=O(1)$. Evidence-tracking memory retains the conditional
bound, while the prompt composer carries these missing justifications as
proof objectives. The research program can then investigate trajectory
control using the local result without assuming those conditions hold.

\textbf{3. Establishing applicability and target coverage.}
Local reasoning derives
$F(x_N)-F^*\le 2\bar L R^2/(N+1)^2$, conditional on a curvature bound
$\bar L$ along the generated trajectory, where $F^*$ is the minimum value
and $N$ counts iterations. Further investigation supplies a trajectory
certificate with $\bar L=e^{1+L_1R}L_{\mathrm{eff}}$ and uses proximal
optimality to control the update segments. These arguments justify applying
the retained convergence bound. The whole prover substitutes the certificate
and combines the conclusion with the other branches. Original-target
review distinguishes this conditional rate from an unrestricted one: the
requested leading order follows when $L_1R$ is bounded by an absolute
constant. This condition remains attached to the final result, whose
contribution is assessed within the alternatives permitted by $T_0$.

\section{ResearchProofBench}
\label{sec:benchmark}

\bench\ evaluates research-level proof workflows in applied mathematics,
with an emphasis on optimization. It comprises two components: known-proof measures proof construction for published results,
and conjecture research assesses conclusions developed for open problems.
Both assess the submitted argument against the original task, including
the assumptions supporting each result and the obligations left unresolved.

\subsection{Task Components and Literature Access}
\label{sec:task-components}
\label{sec:search-conditions}
\label{sec:conjecture-task}
\textbf{Known-proof.}
The known-proof component comprises 51 proof-reconstruction tasks from 24
published papers in applied mathematics, with 331 evaluator checkpoints
covering registered mathematical claims and their dependencies. Source
selection prioritizes recent papers. Difficulty levels S1/S2/S3 contain
11/29/11 tasks. Appendix~\ref{app:benchmark} gives the selection rules and
task inventory. This component specifies frozen clean-web access: retrieved arXiv
versions must have verified release dates before the task cutoff. Agents
receive the task statement and a shared search tool returning titles,
abstracts and metadata from admitted historical versions. Reference proofs,
marking schemes and other evaluator materials remain hidden. Proof
construction uses these statements and abstracts without access to the
retrieved papers' full texts.

\textbf{Conjecture research.}
The conjecture research component comprises six open-problem cases:
C1, accelerated upper and matching lower bounds for composite $(L_0,L_1)$
optimization under two subgradient regimes;
C2 (R22), optimal nonblocking stochastic strongly convex complexity;
C3 (R03), global-condition-number rates for decentralized optimization;
C4 (R45), single-loop $O(1/T)$ stationarity in communication rounds on
time-varying directed networks;
C5 (R60), necessity of the extra $\ln n$ factor in row-stochastic complexity;
and C6, the least Q-order and attainment of exact-line-search BFGS.
Cases C2--C6 are drawn from the OptBound research problem
collection~\citep{reaslab2026optbound}.
Appendices~\ref{app:conjecture-sources} and~\ref{app:conjecture-questions}
give provenance and objectives.
Each task defines a research objective while allowing candidate statements and proof strategies to evolve. Evaluation assesses the mathematical validity
of the final claim and their contribution to the original objective,
including substantive partial progress. Solutions must respect the task
statement and establish nondegenerate mathematical results. Current public
literature is available for related results and proof techniques.

\subsection{Scoring and Independent Review}
\label{sec:scoring}
\textbf{Known-proof.}
An independent LLM reviewer grades the mathematical content of the mathematical claims on a 0--4 scale. Credit follows the
submitted derivation, including equivalent formulations and accepted
alternative arguments for these claims. The task's proof score is
$Q_t=\max\{0,\sum_j a_{tj}-\sum_{\ell\in A_t}d_\ell\}$,
where $a_{tj}$ is checkpoint $j$'s weighted contribution after dependency
adjustment and $A_t$ contains additional gaps with deductions $d_\ell$.
Weighted obligation completion measures the share of obligation weight
fully established. Complete proof rate (CPR) assesses whether the submitted
argument proves the original theorems. Appendix~\ref{app:rubric}
defines the theorem-level decision rule alongside claim-level grading.

\textbf{Conjecture research.}
We adapt the problem-specific 0--7 grading approach of
\citet{ma2026proofbench}, using a fixed scheme to assess each final submission
against the original research objective. Scores distinguish
limited correct content, substantive partial progress and complete
resolution endpoints. Independent reviews identify the
strongest established result, its conditions and the principal unresolved
gap. For task $t$, LLM and human scores
$(s_t^{\mathrm{LLM}},s_t^{\mathrm{Human}})\in\{0,\ldots,7\}^{2}$
are reported separately. Both use
the same scheme, with credit determined by mathematical contribution
rather than a prescribed proof route.
Appendix~\ref{app:open} specifies the score bands, task-specific criteria
and review protocol.

\section{Experiments}
\label{sec:exp-quality}

\subsection{Experimental Setup}
For known-proof, direct prompting, self-check prompting,
QED~\citep{an2026qed}, Rethlas~\citep{ju2026re},
Iteris~\citep{chen2026iteris} and \agent\ use gpt-5.6-sol xhigh
for proof generation, with resource accounting reported in
Appendix~\ref{app:exp-protocol}. Prompting baselines follow
Appendix~\ref{app:baseline-prompts}; QED, Rethlas, Iteris and \agent\ retain
their native workflows and stopping rules. Each agent
baseline contributes the final result delivered under its own workflow's
submission rules. All systems use frozen clean-web
(Section~\ref{sec:search-conditions}). Independent gpt-6-astra xhigh review
assesses each submission and its included references under the same task
rubric, with system identity and internal acceptance labels withheld.

For conjecture research, all three methods use gpt-5.6-sol on the
same six tasks; archived reasoning settings are reported in
Appendix~\ref{app:exp-protocol}. The two baselines and
\agent\ receive the same complete task statements and background materials,
with access to the same permitted information sources. Resource totals
and submission boundaries are reported in Appendix~\ref{app:exp-protocol}. Direct solving stops after one answer; iterative
solving returns its fifth answer after four critique--revision cycles.
\agent\ allocates its budget across research and proof-development roles
and returns the result selected by its stopping rule. Each workflow's
prescribed final output receives gpt-6-astra review under the 0--7 scheme
in Section~\ref{sec:scoring}.
Baseline prompts and evaluation records appear in
Appendices~\ref{app:exp-protocol} and~\ref{app:conjecture-records}.

\subsection{Known-proof}
Table~\ref{tab:direct} compares the six systems' delivered results on
51 known-proof tasks using the three metrics in Section~\ref{sec:scoring}.
Appendix~\ref{app:rubric} provides per-task scores and reproducibility records.

\begin{table}[htbp]
\centering\small\setlength{\tabcolsep}{3pt}
\caption{Known-proof results on 51 tasks per method: component-level scores and theorem-level completion.}
\label{tab:direct}\label{tab:exp-direct}
\begin{tabular}{@{}>{\raggedright\arraybackslash}p{0.20\linewidth}
>{\centering\arraybackslash}p{0.13\linewidth}
>{\centering\arraybackslash}p{0.37\linewidth}
>{\centering\arraybackslash}p{\dimexpr0.30\linewidth-6\tabcolsep\relax}@{}}
\toprule
System & Proof Score & Weighted Obligation Completion (\%) & CPR (\%)\\
\midrule
direct prompting & 59.25 & 41.02 & 21.57\\
self-check prompting & 60.87 & 44.08 & 27.45\\
QED & 62.22 & 41.49 & 17.65\\
Rethlas & 67.33 & 49.37 & 31.37\\
Iteris & 68.05 & 50.78 & 33.33\\
\textbf{\agent} & \textbf{75.16} & \textbf{60.75} & \textbf{39.22}\\
\bottomrule
\end{tabular}
\end{table}
\FloatBarrier

\agent\ has the highest observed aggregate Proof Score, weighted
obligation completion and CPR in this 51-task comparison under the systems'
native execution protocols.
The higher obligation completion indicates that the score advantage is
accompanied by a larger share of fully established proof components.
At the theorem level, \agent\ completes 20 of the 51 tasks, compared with
17 for Iteris, yielding CPRs of 39.22\% and 33.33\%, respectively. These metrics describe complementary aspects of the delivered proofs.
Proof Score credits graded mathematical progress after dependency
adjustment and gap deductions, while weighted obligation completion
measures the weight of registered components established in full.
CPR assesses completion of the original theorem along the submitted proof
route, including valid alternative arguments that bypass registered
intermediate claims.
Difficulty-stratified results and paired completion counts appear in
Appendix~\ref{app:paired-completion}.

\subsection{Conjecture-Research Results}

We evaluate the three workflows on the six open-problem cases introduced
in Section~\ref{sec:conjecture-task}, where candidate statements and proof
strategies evolve toward the original research objective. Table~\ref{tab:exp-conjecture} compares single-answer solving, generic critique--revision and \agent's organized research workflow.
Following Section~\ref{sec:scoring}, separate LLM and human assessments
use the same 0--7 scoring scheme to evaluate mathematical validity
and contribution to the original objective, including substantive partial
progress. The table reports scores, established conclusions and remaining
gaps for each final submission.

\begin{table}[htbp]
\centering\small\setlength{\tabcolsep}{3pt}
\caption{Conjecture-research outcomes with LLM and human scores on the 0--7 scale.}
\label{tab:exp-conjecture}
\begin{tabularx}{\linewidth}{@{}l
  >{\raggedright\arraybackslash}p{0.16\linewidth}
  ccYY@{}}
\toprule
ID & Submission & \shortstack{LLM} & \shortstack{Human} & Conclusion & Main gap or qualification\\
\midrule
C1 & Direct Solving & 2 & 2 &
Conditional smoothness reduction &
Invalid refutations\\
\addlinespace[2pt]
 & Iterative Solving & 3 & 3 &
Conditional upper-bound progress &
Initial objective gap omitted\\
\addlinespace[2pt]
 & \agent & 7 & 6 &
Task-permitted conditional result &
Local exposition gap\\
\midrule
C2 & Direct Solving & 2 & 1 &
Sample normalization &
Minimax proof absent\\
\addlinespace[2pt]
 & Iterative Solving & 2 & 2 &
Source normalization &
Minimax resolution absent\\
\addlinespace[2pt]
 & \agent & 5 & 5 &
Fixed-path span lower bound &
Source certification missing\\
\midrule
C3 & Direct Solving & 2 &  2 &
Conditional exact-averaging rate &
Original oracle bridge absent\\
\addlinespace[2pt]
 & Iterative Solving & 2 & 3 &
Conditional oracle reduction &
Gossip cost unproved\\
\addlinespace[2pt]
 & \agent & 4 & 5 &
Global-curvature routed rate &
Spectral target unresolved\\
\midrule
C4 & Direct Solving & 2 & 2 &
Known communication rate &
Single-loop endpoint open\\
\addlinespace[2pt]
 & Iterative Solving & 2 & 2 &
Known communication rate &
Single-loop endpoint open\\
\addlinespace[2pt]
 & \agent & 6 & 6 &
Essentially $O(1/T)$ stationarity &
Startup and output conventions\\
\midrule
C5 & Direct Solving & 1 & 1 &
Comparison issues identified &
No proved partial theorem\\
\addlinespace[2pt]
 & Iterative Solving & 2 & 2 &
Algebraic regime comparisons &
Common-model theorem absent\\
\addlinespace[2pt]
 & \agent & 3 & 4 &
Lower-bound construction &
Extra $\ln n$ gap unresolved\\
\midrule
C6 & Direct Solving & 4 & 4 &
Universal $\rho_n\ge1$ &
Sharpness and attainment open\\
\addlinespace[2pt]
 & Iterative Solving & 3 & 3 &
Orbit order, dimension lift &
Sharp construction missing\\
\addlinespace[2pt]
 & \agent & 7 & 7 &
$\rho_n=1$ attained, $n\ge2$ &
None\\
\bottomrule
\end{tabularx}
\end{table}

\agent\ receives higher scores than direct and iterative solving on all
six cases under both LLM and human review (Table~\ref{tab:exp-conjecture}),
with broadly similar trends across the two assessments. Its outcomes
include conditional conclusions for composite optimization and bounds and
rates for decentralized optimization. For C6, a single infinite exact-BFGS
run attains Q-order one on a globally $C^\infty$, strongly convex objective
in every fixed dimension $n\ge2$ (Appendix~\ref{app:c6-theorem}).
The comparison concerns the mathematical results delivered by each
workflow: established conclusions, justified applicability conditions and
coverage of the original research objective. The remaining gaps distinguish
substantive partial progress from complete resolution.
For C1, \agent\ derives a task-permitted conditional accelerated rate under
explicit applicability conditions; the expert assessment is detailed in
Appendix~\ref{app:c1-expert-assessment}.

\subsection{Workflow and Search Ablations}
\paragraph{Research workflow ablations.}
Table~\ref{tab:components} compares the full system with five workflow
ablations on C1. Single-pass research--proof examines the value
of iterative research and candidate refinement. Fixed-prompt context
organization tests task-specific composition of the same available
evidence. A predetermined research sequence tests adaptive
selection of research activities based on current objectives and feedback. Proof critique and revision examines RCI's joint refinement of
candidate statements and supporting mechanisms. Removing the
revision-triggered revalidation module tests dependency-guided review
of evidence applicability.
All submissions receive anonymous gpt-6-astra review and separate human
assessment under the same C1 0--7 rubric. Intervention details, resource
accounting and submission rules appear in Appendix~\ref{app:exp-protocol}.

\begin{table}[htbp]
\centering\small\setlength{\tabcolsep}{5pt}
\caption{Five research workflow ablations on C1, with the full system as reference.}
\label{tab:components}\label{tab:exp-components}
\begin{tabularx}{\linewidth}{@{}Ycc@{}}
\toprule
Variant & LLM score & Human score\\
\midrule
Full system & 7 & 6\\
Single-Pass Research--Proof & 3 & 3\\
Fixed-prompt workflow & 6 & 5\\
Predetermined research sequence & 6 & 4\\
Proof critique and revision & 5 & 3\\
Without the revision-triggered revalidation module & 3 & 2\\
\bottomrule
\end{tabularx}
\end{table}
\FloatBarrier

The full system receives the highest scores across the C1 workflow ablations
(Table~\ref{tab:components}). Without the revision-triggered revalidation
module, the LLM and human scores are 3 and 2, compared with 7 and 6 for
the full system.
The intervention details and analysis appear in
Appendices~\ref{app:a5-design} and~\ref{app:ablation-interpretation}.

\paragraph{Search-Condition Ablations.}
Table~\ref{tab:exp-search} compares no-web and live full-web with the
benchmark's frozen clean-web protocol (Section~\ref{sec:search-conditions}).
The comparison assesses how literature availability affects known-proof
performance while holding the full-system architecture, model settings,
tasks and scoring fixed.

\begin{table}[htbp]
\centering\small\setlength{\tabcolsep}{2pt}
\caption{Full-system search ablations with architecture fixed.}
\label{tab:exp-search}
\begin{tabularx}{\linewidth}{@{}Yccc@{}}
\toprule
Access & \mbox{Proof Score} & \mbox{Weighted Obligation Completion (\%)} & \mbox{Complete Proof Rate (\%)}\\
\midrule
No-web & 70.28 & 54.55 & 35.29\\
Frozen clean-web & 75.16 & 60.75 & 39.22\\
Live full-web & 80.82 & 68.90 & 56.86\\
\bottomrule
\end{tabularx}
\end{table}
Frozen clean-web improves all three metrics over no-web, consistent with
the usefulness of historical literature abstracts for constructing and
completing proofs. Live full-web yields the highest scores, but its sources
can include the published solutions themselves, introducing the possibility
of answer retrieval. The frozen condition therefore retains access to
abstracts describing prior results while excluding retrieved versions released after
the cutoff. It serves as the common literature control for the main
comparison, allowing proof development to use existing mathematical
knowledge under the benchmark's historical access rule.

\section{Conclusion}
\label{sec:conclusion}
We introduced \agent\ for research-level proof workflows in which candidate
statements and proof strategies evolve together. Its design separates a result's
validity from its applicability to the current candidate and its contribution
to the original question. Versioned support assessments preserve useful
local results while reopening the conditions affected by revision; RCI
turns these conditions into further candidate refinement and proof objectives.
On \bench's 51 known-proof tasks and six open-problem cases in applied
mathematics, predominantly optimization, \agent\ records the highest
aggregate Proof Score in the known-proof comparison and higher
conjecture-review scores than direct and iterative solving under the
reported workflow protocols. The recorded
investigations show how support reassessment guides continued proof
development across evolving candidates.

\section*{AI Use Statement}
Generative AI tools, including OpenAI Codex, were used to support the research
and writing process. They assisted with formulating and refining mathematical
subproblems and proof obligations, exploring proof strategies, planning
evaluation protocols, searching and summarizing literature, and editing the
manuscript, figures, references, and LaTeX. The authors reviewed AI-assisted
material, verified citations and mathematical claims against original sources
and evaluation records, and tested code where applicable. The runtime LLM
calls of \agent\ are part of the evaluated system and are documented in the
method and evaluation protocols; the authors take responsibility for the final
text, claims, results, citations, and artifacts.

\bibliographystyle{plainnat}
\bibliography{references}
\clearpage
\appendix
\section{Benchmark Construction and Task Inventory}
\label{app:benchmark}
This appendix describes the source selection, mathematical scope and task
inventory of \bench. Known-proof scoring is specified in
Appendix~\ref{app:rubric} and open-case review in Appendix~\ref{app:open}.

\paragraph{Method-independent benchmark design.}
Task selection, task requirements and scoring criteria were developed
independently of \agent's workflows, prompts and skill catalog. They were
not adjusted in response to \agent's outputs or performance. Known-proof
tasks take their mathematical objectives from the selected source results;
conjecture cases take theirs from the original research questions. These
objectives and evaluation criteria apply to every compared system.

\subsection{Source Selection and Mathematical Scope}
The known-proof component contains 51 tasks from 24 papers.
The collection includes sources from SIAM Journal on Optimization, Mathematical
Programming, IMA Journal of Numerical Analysis, Numerische Mathematik and
Mathematics of Computation. Table~\ref{tab:clusters} summarizes the coverage.
Source selection favors recent papers within this mathematical scope.

Each task counts one distinct original result, identified by source paper,
result kind and result number. The 51 results comprise 28 theorems, 20 lemmas,
and 3 propositions. Multiple parameter regimes, proof directions and subclaims
of the same source result are counted once. Long statements, interacting conditions and comparisons between stronger and
weaker assumptions and conclusions motivate this computational-mathematics scope.
Some tasks select an explicit subconclusion or narrower assumptions;
the task statement specifies the selected scope.

Each standalone task statement supplies the necessary definitions and
permitted assumptions. Source titles, theorem locators, reference proofs,
proof obligations and scoring notes are available to evaluators and remain
hidden from the systems during proof generation.

\begin{table}[htbp]
\centering\small
\caption{Source-paper clusters in the known-proof component.}
\label{tab:clusters}
\begin{tabularx}{\linewidth}{llY}
\toprule
Cluster & Tasks & Main mathematical content\\
\midrule
P01 & 1 & Recurrence comparison and nonasymptotic bounds\\
P02 & 2 & Regularized quasi-Newton complexity\\
P04 & 3 & Interpolation, stepsize sequences and sharpness\\
P06 & 2 & Exact-penalty subproblems and duality\\
P08 & 2 & Local subgradient geometry and smoothing\\
P10 & 4 & Primal--dual fixed points, identification and restart\\
P12 & 3 & Constrained stochastic optimization and variance recurrences\\
P13 & 3 & Semismoothness and generalized derivatives\\
P14 & 2 & Convex-cone angles and geometric distances\\
P15 & 2 & Noisy interior-point methods and Newton errors\\
P16 & 2 & Nonconvex subdifferentials and variational convexity\\
P17 & 3 & Minimax constraint qualifications and subproblem well-posedness\\
P18 & 2 & Optimal fixed-point algorithms: necessity and uniqueness\\
P19 & 1 & Truncated moments and SDP exactness\\
P21 & 1 & Algebraic equivalence of oracle algorithms\\
P22 & 2 & Bilevel dynamics and energy estimates\\
P23 & 3 & Cubic--quartic regularization and SDP duality\\
P25 & 2 & Structured matrix perturbations and variable projection\\
P26 & 2 & Primal--dual bounds for heterogeneous compositions\\
P29 & 2 & Wave-equation reconstruction and mesh-change errors\\
P30 & 3 & Constrained Galerkin subproblems and convergence\\
P31 & 2 & Ritz subspaces and certified eigenvalue bounds\\
P32 & 1 & Decentralized primal--dual linear convergence\\
P33 & 1 & Accelerated proximal backtracking\\
\bottomrule
\end{tabularx}
\end{table}

\subsection{Statement Preparation and Hidden Obligations}
Statement preparation specifies the selected source result's mathematical scope:
objects, quantifiers, admissible parameter ranges, algorithm definitions,
and requested conclusions. Notation can be standardized, but the process
must not silently replace an existential conclusion by a universal one,
drop a branch condition, or substitute an easier asymptotic statement for
a quantitative bound. A task that relies on a previous source theorem must
either expose its necessary mathematical content lawfully or require the
agent to obtain an admissible source and justify the transfer.

The hidden obligations specify 331 checkpoints covering registered claims and dependencies,
with 6--17 checkpoints per task and positive integer weights summing to 100.
Each checkpoint has three authored claims: supporting C1--C2 and central C3.
The obligations specify the hypotheses, branches and dependencies needed
for the selected result. A reviewer assesses the candidate's actual argument on the 0--4 scale in
Appendix~\ref{app:rubric}; the scorer propagates prerequisite restrictions and
computes weighted points and independent additional deductions.

Registered equivalence rules admit mathematically equivalent formulations
and task-permitted alternative derivations of individual obligations. Review checks the
content established under the task's hypotheses, including justified changes
of variables and algebraic representations. Each judgment cites the submitted
argument and explains the relevant equivalence, as specified in
Appendix~\ref{app:rubric}.

\subsection{Task Labels and Difficulty}
Tasks are characterized by mathematical domain, topic and the reasoning
required in the proof.
Examples include recurrence comparison, sharpness constructions, duality and
spectral or discretization error bounds.
Attribute groups can overlap: a task may require both a construction and a
sharpness certificate. S1 denotes a local derivation, certificate, or short
key lemma (11 tasks); S2 a theorem proof with several connected steps
(29 tasks); S3 an exact bound, worst-case or end-to-end complexity argument,
long dependency chain, or new construction (11 tasks).

The labels support disaggregated evaluation by domain, mathematical area,
proof attribute and difficulty.
Tables~\ref{tab:audit-one} and~\ref{tab:audit-two} list task identifiers,
source result numbers, difficulty and hidden-obligation counts. A task may
select a stated part of the named source result.

\begin{table}[htbp]
\centering\small
\caption{Known-Proof task inventory (1/2).}
\label{tab:audit-one}
\begin{tabularx}{\linewidth}{@{}lYrr@{}}
\toprule
Task & Source result & Difficulty & Checkpoints\\
\midrule
P01-T01 & Theorem 1 & S2 & 6\\

P02-T02 & Lemma 19 & S1 & 7\\
P02-T03 & Theorem 22 & S2 & 7\\
P04-T04 & Proposition 2.17 & S2 & 6\\
P04-T05 & Lemma 4.3 & S2 & 6\\
P04-T06 & Lemma 5.4 & S2 & 6\\
P06-T04 & Theorem 13 & S2 & 7\\
P06-T05 & Theorem 14 & S2 & 6\\

P08-T05 & Lemma 3.6 & S2 & 6\\
P08-T06 & Lemma 3.8 & S2 & 6\\
P10-T01 & Theorem 1 & S3 & 7\\
P10-T02 & Theorem 3 & S3 & 8\\
P10-T03 & Theorem 4 & S2 & 6\\
P10-T04 & Theorem 7 & S2 & 6\\
P12-T01 & Lemma 1 & S1 & 6\\
P12-T02 & Lemma 4 & S1 & 6\\
P12-T03 & Lemma 5 & S2 & 6\\
P13-T01 & Lemma 3.4 & S1 & 6\\
P13-T02 & Lemma 3.6 & S1 & 6\\
P13-T03 & Proposition 3.8 & S3 & 6\\
P14-T01 & Theorem 5 & S2 & 6\\
P14-T02 & Theorem 13 & S2 & 6\\
P15-T01 & Lemma 3.6 & S2 & 6\\
P15-T02 & Lemma 4.7 & S3 & 6\\
\bottomrule
\end{tabularx}
\end{table}

\begin{table}[htbp]
\centering\small
\caption{Known-Proof task inventory (2/2).}
\label{tab:audit-two}
\begin{tabularx}{\linewidth}{@{}lYrr@{}}
\toprule
Task & Source result & Difficulty & Checkpoints\\
\midrule
P16-T01 & Proposition 7.1 & S2 & 6\\
P16-T02 & Theorem 7.3 & S2 & 6\\
P17-T01 & Theorem 3 & S2 & 6\\
P17-T02 & Theorem 4 & S2 & 7\\
P17-T03 & Lemma 3 & S1 & 6\\
P18-T01 & Theorem 4 & S3 & 6\\
P18-T02 & Theorem 7 & S3 & 6\\
P19-T01 & Theorem 3.1 & S3 & 7\\
P21-T01 & Theorem 5 & S1 & 6\\
P22-T01 & Lemma 2.1 & S2 & 6\\
P22-T02 & Lemma 3.1 & S3 & 6\\
P23-T01 & Theorem 3.2 & S2 & 6\\
P23-T02 & Theorem 4.1 & S3 & 6\\
P23-T03 & Theorem 4.3 & S2 & 6\\
P25-T01 & Theorem 2.6 & S1 & 6\\
P25-T02 & Theorem 2.10 & S2 & 6\\
P26-T01 & Lemma 3.2 & S2 & 6\\
P26-T02 & Lemma 5.2 & S1 & 6\\
P29-T01 & Lemma 4.3 & S2 & 6\\
P29-T02 & Lemma 4.7 & S1 & 6\\
P30-T01 & Theorem 4.2 & S2 & 6\\
P30-T02 & Lemma 5.2 & S2 & 6\\
P30-T03 & Theorem 5.3 & S3 & 7\\
P31-T01 & Theorem 2.6 & S1 & 6\\
P31-T02 & Theorem 3.1 & S2 & 6\\
P33-T01 & Theorem 4.4 & S2 & 11\\
P32-T01 & Theorem 1 & S3 & 17\\
\bottomrule
\end{tabularx}
\end{table}

\subsection{Examples of Condition-Sensitive Evaluation}
The generalized Chung lemma task, P01-T01, illustrates why citing a familiar
name is not sufficient. Its hidden reference separates the changing envelope
from contraction of the excess above that envelope. In this example,
$r:I\to\mathbb{R}$ is the task's convex envelope function on a real interval
$I$, $b_k\in I$ is its evaluation sequence and $k$ indexes recurrence steps.
Writing $\partial r(b_k)$ for the convex subdifferential, a subgradient
$g_k\in\partial r(b_k)$ gives the convexity step
$r(b_{k+1})-r(b_k)\geq g_k(b_{k+1}-b_k)$; the argument then relates envelope
variation to the recurrence and extends the bound over the requested
index range. Merely invoking the requested lemma does not establish it.
An independently established, permitted result is assessed as source use,
with applicability justified and retrieval contribution reported separately.
Equivalent algebraic presentations of the recurrence argument receive credit
when they establish the registered claims with the required signs and index ranges.

The finite-identification task P10-T03 makes the identity of the sequence
part of the target. Here $j$ indexes iterates, $w^j=(x^j,y^j)$ contains
the primal and dual components and $F$ is the task-specified primal--dual
mapping. The shadow points are
$\widehat w^j=F(w^j)=(\widehat x^j,\widehat y^j)$; both sequences are
assumed to converge to a solution $w^*=(x^*,y^*)$.
Under partial smoothness and the
stated nondegeneracy condition, the requested conclusion is eventual
membership of the primal shadow component $\widehat x^j$ in the active
manifold at $x^*$. It does not request
finite identification by the anchors $x^j$. A proof about the wrong sequence
does not become acceptable because its terminology and convergence argument
look plausible.

\section{Knowledge Access and Run Isolation}
\label{app:knowledge}
\subsection{Frozen literature access}
The common frozen clean-web protocol applies the same literature-admission
criterion to every system. Let $\mathcal{D}^{\mathrm{arXiv}}$ denote arXiv
versions with verified public release dates and $\kappa_{p(t)}$ the exclusive
cutoff for the source paper of task $t$. All tasks from that paper share this
cutoff. Writing $\operatorname{public}(d)$ for the release date of version $d$,
the admissible literature is
\begin{equation}
 \mathcal{D}_t^{\mathrm{allowed}}
 =\{d\in\mathcal{D}^{\mathrm{arXiv}}:
       \operatorname{public}(d)<\kappa_{p(t)}\}.
 \label{eq:frozen-set}
\end{equation}
The cutoff is the earliest public-answer date documented in the source
registry, or an earlier conservative date. A later reference revision does
not move it forward. Date-only cutoffs are interpreted as 00:00 UTC, excluding
that entire day.

For each retrieved paper, search selects the latest verified version strictly
before the cutoff and returns its title, historical abstract, release date,
version-specific source link and content hash. During generation, the model
sees the task statement and these search results. The abstract is the only
paper text supplied by the tool: it returns no body excerpts or full theorem
or proof text beyond material appearing in the abstract itself. Source links
identify the historical versions and do not permit separate document downloads.
The protocol provides no preloaded background corpus and exposes only the
search tool; direct browsing and unrestricted retrieval are excluded. An empty search or service failure does
not permit fallback to a current version or live web search.

\subsection{Gateway Enforcement and Coverage}
The generation package contains 51 standalone statements, 24 paper-cutoff
policies, generation runners and one authenticated MCP tool,
\texttt{search(query, top\_k)}. Each agent receives only its own statement,
approved static code and a private workspace. The 24 target-source PDFs,
references, marking schemes and review records remain evaluator-side.
The host binds a private token to each task; query arguments cannot change
its task, paper, or cutoff. Under the unified protocol, the same search tool
and document-admission rules apply to all systems, with no method-specific
search-call restriction.

The gateway stores hash-checked responses and exact query results in a private
run cache, with a hash-chained audit log. Repeating a query replays its retained
result; replay-only mode rejects uncached queries and makes no new network
requests.

The MCP broker runs outside the candidate workspace. Filesystem and network
isolation must deny access to other tasks, evaluator materials, raw caches,
direct browsing or downloads and provider-hosted retrieval. No \texttt{open}
tool is exposed. Dedicated workers avoid shared temporary-directory leakage.
The production adapter is checked for tool exposure, task-token isolation,
cutoff enforcement, cache integrity and logged retrieval; a condition that
cannot be enforced is marked unavailable. These access checks are separate
from mathematical review.

\subsection{Statement-to-Proof Generation}
The generation runner reads the pinned task statement and starts an isolated
attempt with the registered prompt and tool configuration. Under the unified
Frozen protocol, direct prompting, self-check prompting and native agents
use the same search tool and literature-access restrictions. The prompting
baselines use the templates in Appendix~\ref{app:baseline-prompts}; native
agents retain their own proof workflows.

A successful attempt retains the final mathematical text together with
the task, generation mode and completion record. Exact prompts, model responses and MCP
and retrieval records identify each submitted proof. Generation preserves
the submitted mathematics without rewriting it; anonymization and independent
review follow as a separate stage.

\subsection{Within-Task Memory and Retrieval Attribution}
Within-task memory is part of the evaluated architecture: agents may retain
plans, failed routes, source records and verification feedback across their
own calls. Each new problem starts with an empty task workspace; fixed general
skills cannot be edited from test outcomes during the comparison.

Retrieval attribution records whether the text returned to the agent supplies
a definition, an applicable result, a mechanism that still needs derivation,
or the requested answer. Under Frozen access, this text is the historical
abstract. Review records identify the matched passage and the bridge supplied
by the agent, separating retrieved content from the proof contribution.

\section{Evaluation Rubrics and Statistical Units}
\label{app:rubric}
\paragraph{Proof obligations and admissible derivations.}
\label{app:score-definitions}
Human domain experts reviewed the scoring scheme's criteria. Each task
has 6--17 checkpoints with positive integer weights summing to 100.
The collection contains 331 checkpoints and 993 authored claims. Each checkpoint
has supporting claims C1--C2 and a central claim C3, each receiving one third of
its weight. The task defines the requested theorem and its assumptions. The rubric
provides a fixed decomposition into supporting and central claims, with
weights and dependencies for component-level assessment.

\paragraph{Mathematical equivalence and accepted derivations.}
Claims receive credit for the mathematical content established by the
submitted argument. Accepted derivations include equivalent formulations
and registered alternatives: P15-T02 accepts the endpoint reciprocal
identity or its matrix form, and P32-T01 accepts a decrement estimate or a
proved graph-based conversion from energy contraction to the required
Euclidean error bound. The reviewer checks the derivation and its
applicability under the task's assumptions.

Six frozen calibration pairs test representation invariance through
notation changes, equivalent identities, nodal interpolation and sign
normalization. Each pair receives identical effective grades and task
scores, including the pair of incomplete arguments.

\begin{table}[htbp]
\centering\small
\caption{Atomic grades for registered mathematical claims.}
\label{tab:obligation-credit}
\begin{tabularx}{\linewidth}{@{}cY@{}}
\toprule
Grade & Required mathematical evidence\\
\midrule
4 & A complete, correct derivation, including hypotheses, quantifiers, constants, signs, branches and theorem applicability; the reviewer need not supply a missing argument.\\
3 & The derivation is correct apart from a minor local rigor gap.\\
2 & Valid progress is present, but a necessary bridge remains unproved.\\
1 & A relevant proof attempt contains a mathematical error or invalid argument.\\
0 & No supporting derivation of the registered claim is supplied.\\
\bottomrule
\end{tabularx}
\end{table}

Each grade has a checkable mathematical justification and cites the
submitted proof or its included references. Grades below 4 identify the
unestablished claim or invalid inference. The reference supplies context
for interpreting the assessment claims.

\paragraph{Dependency adjustment and task score.}
The registered claim-dependency graph is acyclic. C3 depends on C1--C2 within
its checkpoint; a registered prerequisite checkpoint contributes all its
claims as dependencies. Let $g_{tjk}$ be the reviewer's raw grade and
$\mathcal D_{tjk}$ the registered prerequisite claims. In dependency order,
\begin{equation}
 \widetilde g_{tjk}
 =\min\!\left(\{g_{tjk}\}\cup
 \{h(\widetilde g_{td}):d\in\mathcal D_{tjk}\}\right),\qquad
 h(q)=\begin{cases}
 2,&q<3,\\
 3,&q=3,\\
 4,&q=4.
 \end{cases}
 \label{eq:grade-dependencies}
\end{equation}
Here $d$ indexes an atomic prerequisite. Each effective grade is computed once
and used by its dependents. With checkpoint weights $w_{tj}>0$ summing to
100 and equal claim shares $\alpha_{tjk}=1/3$, the checkpoint contribution is
$a_{tj}=w_{tj}\sum_{k=1}^{3}\alpha_{tjk}\widetilde g_{tjk}/4$.

Each gap has a unique issue identifier, description, severity, affected claims,
and evidence locations. A gap already reflected in claim grades cannot also
receive an additional deduction. Let $A_t$ contain the distinct additional
gaps, with deductions $d_\ell\in\{3,10,30\}$ for minor, major and fatal
severity, respectively. The task's proof score is
$Q_t=\max\{0,\sum_j w_{tj}\sum_{k=1}^{3}\alpha_{tjk}\widetilde g_{tjk}/4-\sum_{\ell\in A_t}d_\ell\}$,
where $j$ ranges over its registered checkpoints. Numerical aggregation uses
exact rational arithmetic before rounding exported scores to six decimal places.

\paragraph{Obligation completion and complete proofs.}
\label{app:woc-definitions}
For each obligation, define
$C_{tj}=\mathbf{1}\{\widetilde g_{tjk}=4\text{ for all }k\}$.
Using effective grades ensures that a missing prerequisite prevents full
obligation credit. The task-level percentage is
$W_t=100(\sum_j w_{tj}C_{tj})/(\sum_j w_{tj})=\sum_j w_{tj}C_{tj}$,
since $\sum_j w_{tj}=100$. The weighted obligation completion averages these percentages across the same
51 tasks used for Proof Score; it does not pool all 331 obligations with
equal weight. A local grade~3 therefore leaves an obligation outside weighted obligation completion,
while its partial credit remains in Proof Score.

\paragraph{Complete proofs through alternative routes.}
\label{app:theorem-completion}
A submission constitutes a complete proof if it establishes the original
task's conclusion under the permitted assumptions through a mathematically
valid and complete argument. Alternative proofs may bypass registered
intermediate claims; unused reference lemmas need not be established.
For example, if the reference proves $A\Rightarrow B\Rightarrow C$, a
submission that establishes $D$ and derives $C$ from $D$ is complete when
both steps are justified under the task's assumptions. Proof Score and
weighted obligation completion retain their component-level interpretation,
while complete-proof status follows the current review's theorem-level
assessment against the original task.

The theorem-level decision uses the mathematical explanations, route
assessment and gap records. To assign $P_t^{\mathrm{thm}}=1$, the review
must establish three findings: the submitted argument reaches the original
target conclusion; its assumptions and invoked results are admissible for
the task; and every inference required by its actual proof route is
justified. An unresolved gap in that route or an unmet task requirement
gives $P_t^{\mathrm{thm}}=0$. Omission of a registered intermediate claim
is relevant to this decision only if the submitted proof depends on it.
The absence of a recorded gap alone is insufficient: the explanations
must establish all three findings. CPR is
$\mathrm{CPR}=100N^{-1}\sum_{t=1}^{N}P_t^{\mathrm{thm}}$.
Claim-level grades determine component credit; the theorem-level findings
determine the CPR label.

\paragraph{Relationship to Proof Score.}
Let $R_t$ be the pre-deduction points earned by obligations that are not fully
complete and let $\Delta_t$ be the actual additional deduction after applying
the zero floor. The decomposition
\[
 Q_t=W_t+R_t-\Delta_t
\]
is exact because each fully completed obligation contributes its entire
weight to the score. Thus weighted obligation completion identifies the contribution from fully
completed proof components; Proof Score also credits partial derivations.
For example, with four independent, equally weighted obligations and no
additional deduction, completing two obligations gives Proof Score~50 and
weighted obligation completion~50\%. Giving every claim grade~2 also gives Proof Score~50, but weighted obligation completion~0\%.
These examples distinguish partial-credit Proof Score from weighted
obligation completion.

\paragraph{Final obligation-completion records.}
\label{app:woc-records}
Table~\ref{tab:woc-summary} lists the final reviewed-task counts,
completed-obligation counts and weighted obligation completion.
The metric follows the definition in Appendix~\ref{app:woc-definitions}.

\begin{table}[htbp]
\centering\small\setlength{\tabcolsep}{4pt}
\caption{Final weighted obligation completion for the same records as Table~\ref{tab:direct}. Obligation counts are descriptive; Weighted Obligation Completion uses task-specific weights and equal averaging over 51 tasks.}
\label{tab:woc-summary}
\begin{tabularx}{\linewidth}{@{}Yccc@{}}
\toprule
System & Reviewed tasks & Complete obligations & \shortstack{Weighted Obligation\\Completion (\%)}\\
\midrule
direct prompting & 51/51 & 146/331 & 41.02\\
self-check prompting & 51/51 & 155/331 & 44.08\\
QED & 51/51 & 152/331 & 41.49\\
Rethlas & 51/51 & 171/331 & 49.37\\
Iteris & 51/51 & 178/331 & 50.78\\
\agent & 51/51 & 206/331 & 60.75\\
\bottomrule
\end{tabularx}
\end{table}

\paragraph{Independent review and evidence traceability.}
\begin{table}[htbp]
\centering\small
\caption{Inputs and outputs of one independent mathematical review.}
\label{tab:review-inputs}
\begin{tabularx}{\linewidth}{@{}lY@{}}
\toprule
Available & Literal task, permitted assumptions, anonymized final proof and referenced proof bodies, hidden reference, registered claims, weights and dependencies\\
Excluded & System identity, internal acceptance labels, other submissions and other review judgments\\
Model output & Atomic grades, route classifications, mathematical explanations, gap records and supporting, contextual, or contradicting line references\\
Derived output & Effective grades, checkpoint points, additional deductions, Proof Score and weighted obligation completion\\
CPR decision & Target conclusion, admissible assumptions and a justified proof route, assessed from the mathematical explanations and gap records\\
\bottomrule
\end{tabularx}
\end{table}
The evaluator uses \texttt{gpt-6-astra} with \texttt{xhigh} reasoning in a fresh
context separate from proof generation. One mathematical review
assesses whether the submission proves the original theorem and grades
every registered claim. Independence refers to the separate context
and withheld system identity. Numerical scores are aggregated from the
claim grades and deductions. The CPR label follows the theorem-level
decision rule in Appendix~\ref{app:theorem-completion}.

The candidate is retained unchanged and presented with stable line identifiers.
Evidence references distinguish support, surrounding context and a faulty
passage. An absence finding may cite neighboring context while describing the
missing inference. The program resolves line references to the original text for traceability.
Claim grades, routes and gap categories must be mutually consistent.
The first complete semantic response is retained. Evidence-only repairs may
correct line references or evidence fields while preserving a semantic
fingerprint of grades, reasons, routes and gaps. A low score or failed
validation does not trigger a new mathematical judgment. Unresolved structure,
location or semantic conflicts retain unknown outcomes. Full-score frequency
is diagnostic only; it neither suspends review nor changes individual grades.

\paragraph{Evaluation and reporting records.}
\label{app:reproduce-table3}
The benchmark fixes the 51 task statements, reference proofs, claim weights,
dependencies, review instructions and scoring code. The reported comparison
contains 306 evaluated submissions and reviews, with each submission linked
to its task, model configuration and retained judgment. The three
reported metrics are computed from these fixed judgments using the definitions
above.
The \agent\ records in Table~\ref{tab:direct} are the same 51 Frozen clean-web
records used in Table~\ref{tab:exp-search}. The search comparison also retains
51 records for each of No-web and Live full-web.

The search-condition replay checks the identities of the 153 retained records
and reproduces all nine cells of Table~\ref{tab:exp-search}. Candidate, rubric and semantic identities
are checked before exact aggregation. Hosted model identifiers are fixed,
while fresh model judgments may vary; replay of retained judgments makes no
model requests.

\paragraph{Final scoring records.}
\label{app:final-scoring-records}
The final inventory contains 306 accepted reviews: all 51 tasks for each of
direct prompting, self-check prompting, QED, Rethlas, Iteris and \agent.
Each task has the same rubric and checkpoint weights across the six methods.
Every record specifies the candidate and rubric identities, raw and effective
claim grades, supporting evidence, mathematical gaps, additional deductions,
Proof Score, weighted obligation completion and theorem-level complete-proof status.
Offline score replay reproduces the scores for all 306 records, checks candidate and rubric
hashes and recomputes effective grades, obligation points and additional
deductions without model calls. Referenced proof bodies accompany each
anonymized candidate; the evaluator does not retrieve citations automatically.
There are no unknown selected outcomes. Means use exact weighted claim scores
before rounding. The metric export is versioned separately from the
mathematical judgments.

The final task contracts explicitly state the permitted assumptions and
background results. P17-T03 uses the lifted set $Y\times Y$ and upper
semicontinuity at a fixed parameter, without simplex variables or continuity
of the common additive term. P14-T02 requires the attaining antipodal point
and the stated aperture range, not full-sphere coverage. In P15-T02, the
endpoint reciprocal identity and its equivalent matrix form establish the
same signed barrier remainder. P30-T02 admits any verified nodal cardinal
interpolant, including midpoints, while retaining the separate coercivity
obligation. For P04-T04, consistent variable renaming and sign reversal
preserve a proof step; they cannot supply an omitted factor-monotonicity
argument.

P17-T01 requires the finite-family fixed-witness, neighborhood and
finite-subcover argument. For P26-T01, coercivity concerns the auxiliary
variable $w$ with $x$ and $\lambda^\star$ fixed. The convex sum rule constructs
$\widehat\lambda\in\partial h(\widehat g)$ with
$\|\widehat\lambda-\lambda^\star\|\le r$; conjugacy establishes its dual
feasibility. The zero-displacement case uses the stationarity-selected norm
subgradient. The first lifted Lagrangian term uses $x$, and the second term
is bounded above by $F(x^\star)$ using the granted saddle inequality or its
registered Fenchel expansion. The residual estimate follows from the saddle
and Fenchel nonnegative terms without requiring $g(x)\in\operatorname{dom}h$.

P32-T01 uses the unique optimizer granted in the task. A consistent incidence
normalization fixes the Gram and spectral identities throughout the energy
argument. Either the decrement estimate or a proved graph-based energy
conversion may yield the Euclidean prefactor, with the initialization and
$k=0$ case covered explicitly. The invariant, range restrictions, spectral
bounds and contraction argument are assessed as separate obligations.
P25-T02 grants the eliminated-value and optimizer identities and includes
$dK=-K(dM\,M^*+M\,dM^*)K$ as an explicit obligation. Granted background
facts need not be reproved.

Tables~\ref{tab:final-task-scores-one} and~\ref{tab:final-task-scores-two}
report the five baselines as Proof Score / weighted obligation completion (\%).
Table~\ref{tab:mathsail-final-task-scores} reports both values for \agent\ on all 51 tasks. The first number includes partial
credit and the existing deductions; the second includes only the weight of
fully completed obligations. A Proof Score of 100.00 indicates that all registered
obligations are fully established and no gap remains. A task with weighted obligation completion~100.00 can still have a lower Proof
Score when an additional retained gap remains.

\FloatBarrier
\begin{table}[htbp]
\centering\small\setlength{\tabcolsep}{3pt}
\caption{Final Known-Proof results (1/2). Each entry is Proof Score / Weighted Obligation Completion (\%), computed from the same accepted review.}
\label{tab:final-task-scores-one}
\begin{tabularx}{\linewidth}{@{}l*{5}{>{\raggedleft\arraybackslash}X}@{}}
\toprule
Task & \shortstack{Direct\\Prompting} & \shortstack{Self-Check\\Prompting} & QED & Rethlas & Iteris\\
\midrule
P01-T01 & $58.67\,/\,38.00$ & $51.67\,/\,12.00$ & $100.00\,/\,100.00$ & $100.00\,/\,100.00$ & $100.00\,/\,100.00$\\
P02-T02 & $70.67\,/\,48.00$ & $64.33\,/\,24.00$ & $53.00\,/\,24.00$ & $53.00\,/\,24.00$ & $100.00\,/\,100.00$\\
P02-T03 & $96.67\,/\,90.00$ & $100.00\,/\,100.00$ & $39.50\,/\,0.00$ & $46.50\,/\,0.00$ & $50.00\,/\,0.00$\\
P04-T04 & $36.00\,/\,12.00$ & $44.33\,/\,12.00$ & $28.33\,/\,0.00$ & $100.00\,/\,100.00$ & $69.00\,/\,38.00$\\
P04-T05 & $39.33\,/\,22.00$ & $45.33\,/\,22.00$ & $29.83\,/\,0.00$ & $47.67\,/\,10.00$ & $47.67\,/\,10.00$\\
P04-T06 & $0.00\,/\,0.00$ & $0.00\,/\,0.00$ & $51.00\,/\,10.00$ & $51.00\,/\,10.00$ & $51.00\,/\,10.00$\\
P06-T04 & $33.67\,/\,12.00$ & $37.33\,/\,12.00$ & $34.00\,/\,12.00$ & $41.33\,/\,12.00$ & $28.00\,/\,12.00$\\
P06-T05 & $38.00\,/\,10.00$ & $38.00\,/\,10.00$ & $55.00\,/\,10.00$ & $48.00\,/\,10.00$ & $44.33\,/\,10.00$\\
P08-T05 & $29.00\,/\,0.00$ & $15.33\,/\,0.00$ & $49.00\,/\,0.00$ & $45.33\,/\,0.00$ & $45.33\,/\,0.00$\\
P08-T06 & $13.00\,/\,0.00$ & $20.33\,/\,0.00$ & $29.83\,/\,0.00$ & $38.17\,/\,0.00$ & $31.67\,/\,0.00$\\
P10-T01 & $37.00\,/\,0.00$ & $37.00\,/\,0.00$ & $51.67\,/\,0.00$ & $84.83\,/\,36.00$ & $100.00\,/\,100.00$\\
P10-T02 & $42.67\,/\,14.00$ & $42.67\,/\,14.00$ & $79.00\,/\,58.00$ & $92.00\,/\,80.00$ & $71.50\,/\,40.00$\\
P10-T03 & $96.67\,/\,90.00$ & $96.67\,/\,90.00$ & $93.50\,/\,66.00$ & $100.00\,/\,100.00$ & $100.00\,/\,100.00$\\
P10-T04 & $100.00\,/\,100.00$ & $100.00\,/\,100.00$ & $68.00\,/\,40.00$ & $97.00\,/\,100.00$ & $100.00\,/\,100.00$\\
P12-T01 & $100.00\,/\,100.00$ & $100.00\,/\,100.00$ & $100.00\,/\,100.00$ & $100.00\,/\,100.00$ & $100.00\,/\,100.00$\\
P12-T02 & $77.00\,/\,62.00$ & $77.00\,/\,62.00$ & $69.00\,/\,62.00$ & $77.00\,/\,62.00$ & $100.00\,/\,100.00$\\
P12-T03 & $33.00\,/\,0.00$ & $32.33\,/\,0.00$ & $55.33\,/\,44.00$ & $48.67\,/\,34.00$ & $80.00\,/\,60.00$\\
P13-T01 & $100.00\,/\,100.00$ & $100.00\,/\,100.00$ & $100.00\,/\,100.00$ & $100.00\,/\,100.00$ & $100.00\,/\,100.00$\\
P13-T02 & $57.67\,/\,34.00$ & $53.00\,/\,34.00$ & $55.67\,/\,22.00$ & $53.33\,/\,22.00$ & $100.00\,/\,100.00$\\
P13-T03 & $32.67\,/\,0.00$ & $43.33\,/\,14.00$ & $42.00\,/\,0.00$ & $49.00\,/\,14.00$ & $38.00\,/\,0.00$\\
P14-T01 & $78.83\,/\,12.00$ & $78.83\,/\,12.00$ & $93.17\,/\,62.00$ & $100.00\,/\,100.00$ & $100.00\,/\,100.00$\\
P14-T02 & $100.00\,/\,100.00$ & $100.00\,/\,100.00$ & $100.00\,/\,100.00$ & $100.00\,/\,100.00$ & $100.00\,/\,100.00$\\
P15-T01 & $79.17\,/\,12.00$ & $100.00\,/\,100.00$ & $56.00\,/\,12.00$ & $55.00\,/\,12.00$ & $53.67\,/\,12.00$\\
P15-T02 & $100.00\,/\,100.00$ & $100.00\,/\,100.00$ & $100.00\,/\,100.00$ & $100.00\,/\,100.00$ & $100.00\,/\,100.00$\\
P16-T01 & $100.00\,/\,100.00$ & $86.00\,/\,36.00$ & $67.00\,/\,60.00$ & $100.00\,/\,100.00$ & $100.00\,/\,100.00$\\
P16-T02 & $45.17\,/\,12.00$ & $45.17\,/\,12.00$ & $48.33\,/\,0.00$ & $41.67\,/\,0.00$ & $47.67\,/\,0.00$\\
\bottomrule
\end{tabularx}
\end{table}

\begin{table}[htbp]
\centering\small\setlength{\tabcolsep}{3pt}
\caption{Final Known-Proof results (2/2). Each entry is Proof Score / Weighted Obligation Completion (\%), computed from the same accepted review.}
\label{tab:final-task-scores-two}
\begin{tabularx}{\linewidth}{@{}l*{5}{>{\raggedleft\arraybackslash}X}@{}}
\toprule
Task & \shortstack{Direct\\Prompting} & \shortstack{Self-Check\\Prompting} & QED & Rethlas & Iteris\\
\midrule
P17-T01 & $65.33\,/\,60.00$ & $65.33\,/\,60.00$ & $45.83\,/\,18.00$ & $100.00\,/\,100.00$ & $65.33\,/\,60.00$\\
P17-T02 & $92.33\,/\,74.00$ & $82.33\,/\,68.00$ & $56.83\,/\,24.00$ & $56.00\,/\,24.00$ & $71.67\,/\,46.00$\\
P17-T03 & $100.00\,/\,100.00$ & $100.00\,/\,100.00$ & $59.67\,/\,18.00$ & $86.83\,/\,42.00$ & $86.83\,/\,42.00$\\
P18-T01 & $19.00\,/\,12.00$ & $19.00\,/\,12.00$ & $27.17\,/\,12.00$ & $23.67\,/\,12.00$ & $25.17\,/\,12.00$\\
P18-T02 & $100.00\,/\,100.00$ & $100.00\,/\,100.00$ & $100.00\,/\,100.00$ & $100.00\,/\,100.00$ & $100.00\,/\,100.00$\\
P19-T01 & $35.00\,/\,12.00$ & $14.33\,/\,0.00$ & $58.00\,/\,58.00$ & $60.67\,/\,58.00$ & $58.00\,/\,58.00$\\
P21-T01 & $100.00\,/\,100.00$ & $100.00\,/\,100.00$ & $100.00\,/\,100.00$ & $100.00\,/\,100.00$ & $100.00\,/\,100.00$\\
P22-T01 & $100.00\,/\,100.00$ & $100.00\,/\,100.00$ & $55.00\,/\,10.00$ & $51.33\,/\,10.00$ & $44.33\,/\,10.00$\\
P22-T02 & $88.83\,/\,50.00$ & $100.00\,/\,100.00$ & $100.00\,/\,100.00$ & $100.00\,/\,100.00$ & $68.33\,/\,30.00$\\
P23-T01 & $18.00\,/\,10.00$ & $32.00\,/\,10.00$ & $50.00\,/\,46.00$ & $54.00\,/\,46.00$ & $54.00\,/\,46.00$\\
P23-T02 & $71.33\,/\,50.00$ & $71.33\,/\,50.00$ & $78.33\,/\,64.00$ & $68.67\,/\,64.00$ & $74.67\,/\,64.00$\\
P23-T03 & $50.00\,/\,44.00$ & $29.33\,/\,16.00$ & $39.00\,/\,44.00$ & $44.67\,/\,36.00$ & $44.00\,/\,44.00$\\
P25-T01 & $63.00\,/\,52.00$ & $90.00\,/\,90.00$ & $47.33\,/\,28.00$ & $90.00\,/\,90.00$ & $90.00\,/\,90.00$\\
P25-T02 & $49.17\,/\,14.00$ & $51.00\,/\,36.00$ & $49.67\,/\,36.00$ & $100.00\,/\,100.00$ & $49.17\,/\,14.00$\\
P26-T01 & $70.33\,/\,36.00$ & $100.00\,/\,100.00$ & $90.00\,/\,100.00$ & $100.00\,/\,100.00$ & $100.00\,/\,100.00$\\
P26-T02 & $27.33\,/\,0.00$ & $27.33\,/\,0.00$ & $88.00\,/\,44.00$ & $34.67\,/\,0.00$ & $51.00\,/\,10.00$\\
P29-T01 & $47.67\,/\,14.00$ & $61.83\,/\,14.00$ & $44.33\,/\,14.00$ & $41.00\,/\,14.00$ & $47.67\,/\,14.00$\\
P29-T02 & $24.00\,/\,0.00$ & $32.00\,/\,12.00$ & $38.00\,/\,12.00$ & $23.00\,/\,0.00$ & $28.00\,/\,0.00$\\
P30-T01 & $18.00\,/\,18.00$ & $18.00\,/\,18.00$ & $6.00\,/\,0.00$ & $6.00\,/\,0.00$ & $9.00\,/\,0.00$\\
P30-T02 & $57.50\,/\,14.00$ & $57.50\,/\,14.00$ & $55.17\,/\,0.00$ & $40.33\,/\,0.00$ & $50.00\,/\,14.00$\\
P30-T03 & $49.33\,/\,38.00$ & $49.33\,/\,38.00$ & $64.00\,/\,60.00$ & $54.50\,/\,22.00$ & $78.00\,/\,60.00$\\
P31-T01 & $14.00\,/\,10.00$ & $20.50\,/\,0.00$ & $27.33\,/\,10.00$ & $22.33\,/\,0.00$ & $14.00\,/\,10.00$\\
P31-T02 & $18.00\,/\,0.00$ & $4.67\,/\,0.00$ & $50.00\,/\,44.00$ & $36.00\,/\,30.00$ & $36.00\,/\,30.00$\\
P32-T01 & $100.00\,/\,100.00$ & $100.00\,/\,100.00$ & $100.00\,/\,100.00$ & $100.00\,/\,100.00$ & $100.00\,/\,100.00$\\
P33-T01 & $49.00\,/\,16.00$ & $69.67\,/\,44.00$ & $95.33\,/\,92.00$ & $70.67\,/\,44.00$ & $67.33\,/\,44.00$\\
\bottomrule
\end{tabularx}
\end{table}

\FloatBarrier

\begin{table}[htbp]
\centering\small\setlength{\tabcolsep}{5pt}
\caption{Final \agent\ results on all 51 Known-Proof tasks. The two panels list disjoint task subsets. Each row reports Proof Score and Weighted Obligation Completion (\%); a score of 100 identifies full completion of the registered obligations with no remaining gap.}
\label{tab:mathsail-final-task-scores}
\begin{tabularx}{\linewidth}{@{}l*{2}{>{\raggedleft\arraybackslash}X}@{\hspace{14pt}}l*{2}{>{\raggedleft\arraybackslash}X}@{}}
\toprule
Task & \shortstack{Proof\\Score} & \shortstack{Weighted\\Obligation\\Completion (\%)} & Task & \shortstack{Proof\\Score} & \shortstack{Weighted\\Obligation\\Completion (\%)}\\
\midrule
P01-T01 & 100.00 & 100.00 & P17-T01 & 69.33 & 60.00\\
P02-T02 & 100.00 & 100.00 & P17-T02 & 89.00 & 64.00\\
P02-T03 & 98.33 & 90.00 & P17-T03 & 81.00 & 58.00\\
P04-T04 & 69.00 & 38.00 & P18-T01 & 30.33 & 12.00\\
P04-T05 & 100.00 & 100.00 & P18-T02 & 100.00 & 100.00\\
P04-T06 & 6.00 & 0.00 & P19-T01 & 56.00 & 46.00\\
P06-T04 & 47.67 & 12.00 & P21-T01 & 100.00 & 100.00\\
P06-T05 & 55.00 & 10.00 & P22-T01 & 97.00 & 100.00\\
P08-T05 & 50.67 & 0.00 & P22-T02 & 100.00 & 100.00\\
P08-T06 & 26.67 & 0.00 & P23-T01 & 78.00 & 56.00\\
P10-T01 & 38.33 & 0.00 & P23-T02 & 82.00 & 64.00\\
P10-T02 & 48.00 & 14.00 & P23-T03 & 52.00 & 44.00\\
P10-T03 & 98.33 & 90.00 & P25-T01 & 90.00 & 90.00\\
P10-T04 & 100.00 & 100.00 & P25-T02 & 100.00 & 100.00\\
P12-T01 & 100.00 & 100.00 & P26-T01 & 100.00 & 100.00\\
P12-T02 & 100.00 & 100.00 & P26-T02 & 100.00 & 100.00\\
P12-T03 & 54.67 & 0.00 & P29-T01 & 53.67 & 14.00\\
P13-T01 & 100.00 & 100.00 & P29-T02 & 38.00 & 12.00\\
P13-T02 & 57.67 & 34.00 & P30-T01 & 20.00 & 18.00\\
P13-T03 & 70.33 & 50.00 & P30-T02 & 59.67 & 14.00\\
P14-T01 & 100.00 & 100.00 & P30-T03 & 100.00 & 100.00\\
P14-T02 & 100.00 & 100.00 & P31-T01 & 28.00 & 0.00\\
P15-T01 & 100.00 & 100.00 & P31-T02 & 60.33 & 44.00\\
P15-T02 & 100.00 & 100.00 & P32-T01 & 100.00 & 100.00\\
P16-T01 & 100.00 & 100.00 & P33-T01 & 76.00 & 52.00\\
P16-T02 & 52.33 & 12.00 &  &  & \\
\midrule
\multicolumn{4}{r}{Mean over all 51 tasks} & 75.16 & 60.75\\
\bottomrule
\end{tabularx}
\end{table}

\paragraph{Difficulty-stratified results.}
The following summaries recompute the three Table~\ref{tab:direct} metrics
from the retained task-level records, grouping the same 51 tasks by the S1--S3
labels in Tables~\ref{tab:audit-one} and~\ref{tab:audit-two}. Each method is
averaged over the tasks in the indicated difficulty class. Complete Proof Rate (CPR) reports the number of theorem-level complete
proofs over the class size in parentheses.
These summaries disaggregate the primary comparison by task difficulty.
\agent\ completes 6/11, 9/29 and 5/11 tasks in S1, S2 and S3, respectively.
Its S1 completion count matches Iteris, and its S2 count matches Rethlas.

\begin{table}[htbp]
\centering\small
\caption{Known-Proof results for S1 tasks (11 tasks).}
\label{tab:knownproof-strat-s1}
\begin{tabularx}{\linewidth}{@{}lrrrr@{}}
\toprule
System & $n$ & Proof Score & \shortstack{Weighted Obligation\\Completion (\%)} & \shortstack{Complete Proof\\Rate (\%)}\\
\midrule
Direct Prompting & 11 & 66.70 & 55.09 & 4/11 (36.36)\\
Self-Check Prompting & 11 & 69.47 & 56.55 & 4/11 (36.36)\\
QED & 11 & 67.09 & 47.27 & 3/11 (27.27)\\
Rethlas & 11 & 67.29 & 49.09 & 3/11 (27.27)\\
Iteris & 11 & 79.08 & 68.36 & 6/11 (54.55)\\
\textbf{PACE} & \textbf{11} & \textbf{81.33} & \textbf{72.18} & \textbf{6/11 (54.55)}\\
\bottomrule
\end{tabularx}
\end{table}

\begin{table}[htbp]
\centering\small
\caption{Known-Proof results for S2 tasks (29 tasks).}
\label{tab:knownproof-strat-s2}
\begin{tabularx}{\linewidth}{@{}lrrrr@{}}
\toprule
System & $n$ & Proof Score & \shortstack{Weighted Obligation\\Completion (\%)} & \shortstack{Complete Proof\\Rate (\%)}\\
\midrule
Direct Prompting & 29 & 55.60 & 34.83 & 4/29 (13.79)\\
Self-Check Prompting & 29 & 57.34 & 37.86 & 6/29 (20.69)\\
QED & 29 & 56.38 & 32.55 & 2/29 (6.90)\\
Rethlas & 29 & 64.15 & 44.55 & 9/29 (31.03)\\
Iteris & 29 & 61.62 & 40.48 & 7/29 (24.14)\\
\textbf{PACE} & \textbf{29} & \textbf{72.89} & \textbf{55.79} & \textbf{9/29 (31.03)}\\
\bottomrule
\end{tabularx}
\end{table}

\begin{table}[htbp]
\centering\small
\caption{Known-Proof results for S3 tasks (11 tasks).}
\label{tab:knownproof-strat-s3}
\begin{tabularx}{\linewidth}{@{}lrrrr@{}}
\toprule
System & $n$ & Proof Score & \shortstack{Weighted Obligation\\Completion (\%)} & \shortstack{Complete Proof\\Rate (\%)}\\
\midrule
Direct Prompting & 11 & 61.44 & 43.27 & 3/11 (27.27)\\
Self-Check Prompting & 11 & 61.54 & 48.00 & 4/11 (36.36)\\
QED & 11 & 72.74 & 59.27 & 4/11 (36.36)\\
Rethlas & 11 & 75.76 & 62.36 & 4/11 (36.36)\\
Iteris & 11 & 73.97 & 60.36 & 4/11 (36.36)\\
\textbf{PACE} & \textbf{11} & \textbf{75.00} & \textbf{62.36} & \textbf{5/11 (45.45)}\\
\bottomrule
\end{tabularx}
\end{table}

\paragraph{Paired completion counts and expert re-examination.}
\label{app:paired-completion}
The PACE--Iteris comparison pairs the retained outputs on the same 51 tasks.
Both systems complete 14 tasks; PACE alone completes 6, Iteris alone
completes 3, and neither completes the remaining 28. The net difference
is 3 completed tasks, or 5.88 percentage points in CPR.
Expert re-examination confirmed the reported scores for the submissions
checked, with no score changes.

\paragraph{Aggregation and missing outcomes.}
Tasks have equal weight in the metrics defined in Section~\ref{sec:scoring}. The primary
comparison uses $N=51$ per method; subset summaries state their own task count.
The executable aggregator releases a full-set mean only when every selected
task has an accepted review. Missing proofs, unstarted or cancelled tasks,
and unresolved reviews retain unknown $Q_t$, $W_t$ and $P_t^{\mathrm{thm}}$. A reviewed
submission containing no relevant proof can receive a mathematical zero;
missing output is not imputed as one. For known score sum $S$ and $U$ unknown
tasks, the full-set score bounds are $[S/N,(S+100U)/N]$; the same form applies
to known weighted obligation completion sums. CPR bounds use the same fixed denominator. Output and review coverage accompany these summaries;
reviewed-only means are labelled as conditional subset statistics. The reporting layer exports score, weighted obligation completion and CPR
using retained mathematical judgments. All three preserve unknown outcomes
under the same fixed-denominator rule.

Comparisons use the registered task set and rubric, with generation and
review identities recorded for each selected submission
(Appendix~\ref{app:reproduce-table3}).

\section{Conjecture-Research Evaluation}
\label{app:open}
\subsection{Problem sources}
\label{app:conjecture-sources}
OptBound~\citep{reaslab2026optbound}, developed by the ReasLab Team, organizes
lower and upper complexity bounds in optimization together with their
assumptions, oracle and network models, target metrics and supporting
literature. This organization exposes missing bounds and conditions that
must be reconciled when comparing results. Cases C2--C6 are drawn from the
OptBound problem materials used in this study. We retain the source-entry
identifiers R22, R03, R45 and R60 for C2, C3, C4 and C5, respectively;
C6 retains the BFGS problem identifier \texttt{1983-opt-4} in its source
record. C1 is the composite generalized-smoothness problem used throughout
our method examples. The following descriptions state each research question
and its completion criteria; the accompanying task records fix the precise
assumptions and requested conclusions.

\subsection{Submissions and Source Access}
Each case compares three final mathematical submissions for its original
request: a direct solving answer, the fifth standalone answer from an
independent generic iterative solver chain and \agent's final proof manuscript.
The \agent\ manuscripts for all six cases were generated by its research--proof
workflow. All three submissions are assessed against the same original task;
intermediate drafts remain context records and are not scored.

Current public literature is available under the actual permissions of each
submission. Record the source passage, publication/version information where
available and access date. Baselines cannot access project proofs, another
system's output, or reviewer materials. Source use and originality are assessed separately through documented
literature checks.

\paragraph{Task specification.}
For each of the six cases C1--C6, the original question fixes the mathematical
objects, assumptions, quantifiers, admissible operations, output and requested
conclusions. The task specification also identifies conditions and alternative
endpoints open to investigation. All three submissions are assessed against
this same question and its permitted background. A conditional theorem,
restricted lower bound, or obstruction can constitute a complete answer when
the question expressly permits that endpoint and the proof meets its
requirements. Other useful local results receive partial credit according to
their contribution to the original objective.

\subsection{Problem-specific 0--7 scoring}
Following the problem-specific grading approach of
\citet[Section~2.1]{ma2026proofbench}, each submission receives one holistic
integer score from 0 to 7 under a scheme fixed for its case. The original
question and permitted background determine the scheme, including complete
solutions, partial-progress anchors and major errors; candidate proofs are
excluded when drafting it. The same scheme applies to all three submissions,
and any mathematically valid route can achieve full credit. The common
score bands below are interpreted through the case-specific criteria that
follow; coverage findings explain the judgment and are not added as points.

\begin{center}\small
\begin{tabularx}{\linewidth}{@{}cY@{}}
\toprule
Score & Mathematical outcome\\
\midrule
0 & No confirmable mathematical result addressing the task.\\
1--2 & Limited correct observations or baseline analysis; the research objective remains unresolved.\\
3--4 & A substantive partial result with an essential part of the requested argument unresolved.\\
5--6 & The core result is established in the required model, with localized, repairable gaps.\\
7 & A complete and correct resolution of an endpoint permitted by the original question.\\
\bottomrule
\end{tabularx}
\end{center}

\subsection{Problem descriptions and completion criteria}
\label{app:conjecture-questions}
\paragraph{C1: Composite $(L_0,L_1)$ optimization.}
For $F=f+h$ with $\|\nabla^2 f(x)\|\le L_0+L_1\|\nabla f(x)\|$,
the question asks whether the oracle complexity has order
$R\sqrt{(L_0+L_1\ell)/\varepsilon}$. Here $R$ bounds the initial distance
to an optimizer and $\ell$ bounds subgradients of $h$ either throughout
the relevant region or only at an optimizer. Review separates these strong
and weak regimes and the upper- and lower-bound questions. A complete answer establishes or refutes the proposed
complexity order, or proves it under explicit additional conditions permitted
by the question. It accounts for gradient and proximal calls, parameter
ranges and the dimension and algorithm quantifiers of lower bounds.
An accelerated construction must justify its trajectory conditions and
convergence-to-complexity conversion. A conditional estimate with an unresolved
trajectory or oracle-cost bridge receives partial credit; a fully justified,
nondegenerate completion of an open condition is eligible for full credit.

\paragraph{C2: Stochastic strongly convex optimization (R22).}
On a fixed network, this problem asks whether an existing stochastic
strongly convex upper bound admits a matching lower bound when each worker
can sample and communicate in the same round. The target combines a
statistical accuracy term with a term determined by conditioning and network
connectivity. Completion combines certification of the source's sample normalization with
an authorized minimax result. The certification distinguishes total samples,
samples per worker and nonblocking rounds, preserving the local-output
criterion. The research endpoint may be a matching lower bound for the full
model, a complete result for an expressly permitted linear-span or
fixed-schedule class, or a rigorous obstruction to the proposed bound.
Each route must retain the fixed mixing matrix, stochastic-oracle assumptions,
parameter ranges and resource counts. Normalization alone is limited
progress; a matching theorem with an unresolved certification step remains
incomplete.

\paragraph{C3: Global-curvature complexity (R03).}
For $F=n^{-1}\sum_i f_i$ over a connected network, the problem asks whether
the complexity factor $\sqrt{\kappa_\ell}$ associated with local functions
can be replaced by $\sqrt{\kappa_g}$ for the global objective, while
retaining the dependence on communication latency and the network eigengap.
This matters when averaging improves conditioning, so that
$\kappa_g<\kappa_\ell$. A complete answer gives a decentralized algorithm attaining the global
condition-number rate, a lower bound excluding that rate in the stated model,
or a theorem proving that an additional assumption is necessary together with
the resulting achievable rate. The proof must distinguish local from global
conditioning and account for computation, communication and the stated
accuracy criterion. Oracle changes and stronger graph or function assumptions
must be explicit. A rigorous conditional reduction, restricted-graph result,
or weaker rate can earn substantive partial credit; a sufficient assumption
alone does not establish its necessity.

\paragraph{C4: Time-varying directed networks (R45).}
For smooth nonconvex objectives, the problem asks whether repeated inner
consensus exchanges can be replaced by one optimization update per round
while achieving
$O(1/T)$ average squared-gradient stationarity. It permits either such a
single-loop algorithm or the stated lower bound for an explicit online
algorithm class, establishing the necessity of the logarithmic overhead. A positive result must use each fresh
row-stochastic matrix under the given connectivity assumptions and prove
the guarantee for the specified arithmetic-average output. Both endpoints
must account for initialization, gradients, communication, persistent state,
messages and output construction. Reproducing the known communication rate
is limited progress; an otherwise complete proof with localized startup or
output-accounting issues is assessed within the 5--6 band.

\paragraph{C5: Row-stochastic logarithmic complexity gap (R60).}
For stochastic nonconvex optimization on a fixed row-stochastic network
with $n$ nodes, the question asks whether the additional $\ln n$ in the
network-dependent upper bound can be removed, or whether a corresponding
logarithmic lower bound is unavoidable. Before comparing rates, a proof must reconcile the
theorems' assumptions, algorithm classes, resource counts and stationarity
or output criteria in a common model. An explicit comparison
must include its additional conditions and proof. Rigorous restricted-regime
results and model-reconciliation lemmas receive partial credit; algebraic
comparison of the displayed rates alone does not establish an optimization
theorem or close the gap.

\paragraph{C6: Least Q-order of BFGS.}
The problem asks how slowly exact-line-search BFGS can converge near a
smooth, strongly convex minimizer and whether its slowest admissible
Q-order is attained. A complete answer determines the infimum Q-order $\rho_n$ and whether it is
attained for every fixed $n\ge2$. This requires a universal lower restriction
and a matching admissible construction or equally decisive determination of
the exact value. The proof must preserve the specified Hessian-form BFGS
updates, global exact line searches, $C^\infty$ smoothness, local strong
convexity, positive definiteness and convergence. Any dimension extension
must preserve these properties and the Q-order of the full sequence.
A universal bound, an orbit-order calculation, or a construction with
unresolved realization or attainment establishes only part of the task.

\subsection{Independent LLM and Human Review}
The marking-scheme-only context in \citet[Section~3.1]{ma2026proofbench}
informs the evaluator input: the original task, fixed scheme, anonymized
submission and supplied cited sources. Each submission receives a fresh
context without other submissions, internal acceptance labels, or human scores.
All six cases use \texttt{gpt-6-astra} with \texttt{xhigh} reasoning.
Each submission receives one independent review without tools.

The output contains \texttt{llm\_score} (integer 0--7), \texttt{evidence}
(location, finding and impact), \texttt{conclusion}, \texttt{main\_gap} and
qualitative \texttt{coverage} entries. Coverage supports a holistic judgment
and is not an additive checklist. Conclusion states the strongest established
result and its conditions; Main gap identifies the principal unresolved step,
or None for a complete answer. The common evaluator instruction is:
\begin{quote}\small
Evaluate the supplied submission against the task and fixed 0--7
marking scheme. Check the exact conclusion, assumptions, quantifiers, source
applications, proof steps and requested complexity. Distinguish a demonstrated
error from an unresolved justification. Apply the scheme's full-credit,
partial-progress, major-error and zero-credit rules equally to every route.
Additional conditions may earn full credit only when they faithfully complete
open parts of the request, are noncircular and nondegenerate and the core
conclusion is completely proved. Return: LLM Score (an integer 0--7); evidence
and reasons with proof/source locations; Conclusion (one sentence giving the
strongest established result and essential conditions); Main gap (one sentence,
or None if complete). Do not infer correctness from authorship or internal
acceptance. Use only the supplied task, scheme, submission and permitted
cited sources. Do not combine this score with a human score.
\end{quote}

The human-review protocol uses two independent assessments under the same
scheme, followed by adjudication of differences. LLM and human scores are
recorded separately with the established claims, assumptions and unresolved
gaps. Each record identifies the task, scheme, submission and evaluator
response. Appendix~\ref{app:exp-protocol} gives the generation prompts
and submission rules.

\subsection{Conjecture-case configurations and records}
\label{app:conjecture-records}
For C1--C6, both baselines and \agent\ use \texttt{gpt-5.6-sol}.
They receive the same complete task statements
and background materials, with access to the same permitted information
sources. Direct solving returns one answer; iterative solving
returns the fifth answer after four critique--revision cycles. Each critic
receives the task and current answer, and each revision also receives its
critique. Each method contributes the output delivered under its own
stopping and submission rules
(Appendix~\ref{app:stopping-rules}). Evaluation uses anonymized submissions
with stable line references.

The schemes are drafted by \texttt{gpt-5.6-sol} with \texttt{xhigh} reasoning
from the tasks without candidate proofs and fixed across
the three submissions. The final collection contains 18 accepted Astra reviews:
one for each of the three submissions on C1--C6. Solver-visible statements remain
separate from evaluator-only material.

The final case scores, conclusions and main gaps are listed in
Table~\ref{tab:exp-conjecture}. The stored review records contain the
corresponding line-referenced evidence and frozen marking schemes.

\subsection{Optimization proofs generated by \agent}
\label{app:bound-proof-packages}
\label{app:c6-theorem}
All six \agent\ submissions in Table~\ref{tab:exp-conjecture} were generated
by the agent. The C1 proof is detailed in
Appendix~\ref{app:l0l1-running-example}; this subsection summarizes the
mathematical results and conditions of C2--C6. Their canonical English
manuscripts are bound to the final result manifests, retaining each task,
proof, resource accounting and source record. Independent review assesses
each manuscript against the complete original research question.

\paragraph{C1: Task completion and expert assessment.}
\label{app:c1-expert-assessment}
The expert assessment accepts the core conditional conclusion and assigns
6/7 for a local, repairable gap in the exposition of the Lipschitz estimate.
The upper bound uses the effective estimate
$\bar L=e^{1+L_1R}(L_0+L_1\ell)$, yielding the requested accelerated order
when $L_1R$ is bounded by an absolute constant, or under the stated
trajectory certificate. The requested repair concerns the presentation and
justification of this quantitative estimate within the proof; it leaves
the core conclusion and its task-permitted conditional scope unchanged.
This local exposition gap places the human assessment in the 5--6 band
of the common scheme and is recorded in Table~\ref{tab:exp-conjecture}.
The unrestricted strong-regime absolute-order upper bound and the
corresponding randomized lower-bound extension remain unresolved.

\paragraph{C2: Strongly convex stochastic sample lower bound (R22).}
On the fixed lazy path, let $\rho=\sin^2(\pi/(2M))$ and $L/\mu\ge4$.
For $0<\epsilon\le\Delta_0/2$, the reported total-sample lower bound is
\[
N=\Omega\!\left(\frac{\sigma^2}{\mu\epsilon}
 +M\sqrt{\frac{L}{\mu\rho}}\log_+\frac{\Delta_0}{\epsilon}\right).
\]
Each worker samples and mixes once per round, so $N=MH$. The guarantee
concerns the maximum local expected function error. Separate finite-chain
and Gaussian location families establish the network and statistical terms,
including zero gradient heterogeneity at the optimum.

\paragraph{C3: Global-curvature acceleration with neighbor routing (R03).}
For an $L$-smooth, $\mu$-strongly convex average objective, exact tree
aggregation and broadcast implement acceleration using ordinary local gradients.
With $\kappa_g=L/\mu$, graph diameter $D_G$ and edge latency $\tau$, the
reported time bound is
$O((1+\tau D_G)\sqrt{\kappa_g}\log_+(\Delta_0/\epsilon))$.
A separate $\Omega(D_G)$ causal-hop lower bound holds even at $\kappa_g=1$.
The protocol permits edge-specific neighbor messages and a public routing tree.

\paragraph{C4: Single-loop optimization on time-varying directed networks (R45).}
For row-stochastic matrices with uniformly positive finite-window products,
the manuscript combines finite-cycle ratio averaging and live gradient tracking.
Under its explicit constant-stepsize conditions, it reports
$T^{-1}\sum_{k<T}\|\nabla f(\bar x_k)\|^2=O(1/T)$ for the arithmetic
average at every communication round, with square-summable disagreement.
Each round uses one fresh mixing matrix, one local gradient query and
$3d+1$ transmitted real numbers per sender.

\paragraph{C5: Perron-centroid stochastic and network lower bound (R60).}
For a constructed primitive row-stochastic matrix $A$, let $\beta_A$ be its
Perron-weighted contraction and $\kappa_A=\max_i\pi_i/\min_i\pi_i$.
For $\beta_A\in[0.01,1-1/n]$ and
$K\ge64\max\{n,\sigma^2/(nL\Delta)\}$, the reported lower bound is
\[
\frac1K\sum_{t=1}^K\mathbb E\|\nabla f(\pi^\top X^{(t)})\|^2
 =\Omega\!\left(\sigma\sqrt{\frac{L\Delta}{nK}}
 +\frac{(1+\log\kappa_A)L\Delta}{(1-\beta_A)K}\right).
\]
The causal linear-span protocol queries before mixing and uses $K$ samples
per node. Masked stochastic zero-chains and directed parity chains supply
separate hard families. As in R22, the minimax combination selects the
dominant family and uses $\max\{u,v\}\ge(u+v)/2$.

\clearpage
\subsubsection*{C6: Least Q-order of BFGS and comparison with prior results}
\paragraph{Task assumptions and exact algorithm.}
Fix an integer $n\ge2$. Let $F\in C^\infty(\mathbb R^n)$,
$x_1\in\mathbb R^n$, and $B_1\in\mathbb S_{++}^n$.
Here $\mathbb S_{++}^n$ denotes the symmetric positive-definite matrices.
The Hessian-form BFGS iteration is
\begin{align}
g_k&=\nabla F(x_k),&
d_k&=-B_k^{-1}g_k,&
x_{k+1}&=x_k+\alpha_kd_k,\nonumber\\
s_k&=x_{k+1}-x_k,&
y_k&=g_{k+1}-g_k,&
\alpha_k&\in\mathop{\rm arg\,min}_{\alpha\in\mathbb R}
                 F(x_k+\alpha d_k),\label{eq:c6-iteration}\\
B_{k+1}
&=B_k-\frac{B_ks_ks_k^\top B_k}{s_k^\top B_ks_k}
       +\frac{y_ky_k^\top}{s_k^\top y_k}.\label{eq:c6-update}
\end{align}
Every selected $\alpha_k$ is finite and attains the global minimum on the
whole affine line. An admissible infinite run has all quantities defined,
$s_k\ne0$, $s_k^\top y_k>0$, and $B_k\succ0$ at every iteration.
It converges to a point $x^\star$ with $\nabla F(x^\star)=0$, and there
exist $r,\mu>0$ such that
\[
 \nabla^2F(x)\succeq\mu I_n
 \quad\text{for every }\|x-x^\star\|\le r.
\]
In particular, $\nabla^2F(x^\star)\succ0$. Convexity is required locally
near the limit; the task permits any finite global line minimizer when
there is a tie. The initial positive-definite matrix and the line-search
steps are part of the admissible run.

\paragraph{Q-order and the attainment question.}
With $e_k=\|x_k-x^\star\|$, define
\begin{equation}
 q(F,x_1,B_1)=
 \sup\{p>0:\ \exists\,C>0,\ k_0,\quad e_{k+1}\le C e_k^p
                         \text{ for all }k\ge k_0\}.
 \label{eq:c6-qorder}
\end{equation}
The constants may depend on the run, but not on $k$; a finitely terminating
run has $q=+\infty$. Write $\mathcal A_n$ for the class of admissible runs
and set
\[
 \rho_n=\inf_{(F,x_1,B_1)\in\mathcal A_n}q(F,x_1,B_1).
\]
This definition measures a power bound between every sufficiently late
pair of adjacent errors. It allows $q=1$ together with Q-superlinear
convergence $e_{k+1}/e_k\to0$.

\begin{theorem}[C6: least order and attainment]
\label{thm:c6}
For every fixed integer $n\ge2$, $\rho_n=1$. Moreover, there exists a
single nonterminating admissible Hessian-form BFGS run in that dimension
with $q(F,x_1,B_1)=1$, using exact global line searches.
The constructed objective can be chosen globally $C^\infty$ and globally
strongly convex, with $x^\star=0$ and $\nabla^2F(0)=I_n$.
\end{theorem}

\paragraph{Universal order-one bound.}
On a sufficiently small closed ball around $x^\star$, continuity of the
Hessian gives an upper bound $LI_n$ alongside the lower bound $\mu I_n$.
For all sufficiently late iterates, Taylor's formula and exact-line-search
descent therefore give
\[
 \frac{\mu}{2}e_{k+1}^2
 \le F(x_{k+1})-F(x^\star)
 \le F(x_k)-F(x^\star)
 \le\frac{L}{2}e_k^2.
\]
Thus $e_{k+1}\le\sqrt{L/\mu}\,e_k$, establishing $q\ge1$ for every
admissible run. The attaining construction below uses a single planar
orbit and a single interpolated objective.

\newpage
\paragraph{Exact planar transitions.}
For an integer $p\ge3$, define
\[
 \Phi_p(x,y)=x^py-(p-1)xy^{p+1},\qquad
 G_p(z)=z+\nabla\Phi_p(z).
\]
For $v=(A,\Xi,Y)$ near $v_0=(1,0,-1)$ and $\varepsilon>0$, set
\begin{align*}
 P_p(\varepsilon;v)&=(\varepsilon^pA+\varepsilon^{p^2-1}\Xi,
                            \varepsilon^{p-1}),\\
 R_p(\varepsilon;v)&=(\varepsilon^pA,\varepsilon^{p^2}Y).
\end{align*}
With $q\in\{p,p+1\}$, $m=p(p-1)$, $w=(A_+,\Xi_+,Y_+)$ and
$\tau=\varepsilon^m E$, the four exact transition equations are
\begin{align}
 G_p(R_p)^\top(R_p-P_p)&=0,\nonumber\\
 [G_p(R_p)-G_p(P_p)]^\top(P_q-R_p)&=0,\nonumber\\
 G_q(P_q)^\top(P_q-R_p)&=0,\label{eq:c6-transition}\\
 [G_q(P_q)-G_p(R_p)]^\top(R_q-P_q)&=0.\nonumber
\end{align}
Here $P_q,R_q$ use $(\tau,w)$. Dividing the first two equations by
$\varepsilon^{p^2+p-1}$ and the last two by
$\varepsilon^{p+mq}$ gives a polynomial system extending to
$\varepsilon=0$. Its limiting equations are
\begin{align*}
 A\Xi+Y+A^p&=0,&
 A[\Xi+pA^{p-1}-(p-1)]+Y&=0,\\
 AA_+E^q+YE^{q-1}&=0,&
 (Y+A^p)E^{q-1}&=0.
\end{align*}
They vanish at $(A,\Xi,Y,E)=(1,0,-1,A_+^{-1})$.
At $A_+=1$, subtracting the first linearized equation from the second
isolates $(p-1)^2\,dA$; the fourth, first and third then determine
$dY,d\Xi,dE$. The Jacobian is invertible. The implicit-function theorem
therefore supplies continuous exact transition maps
$(v,E)=S_{p,q}(\varepsilon,w)$, with $E\in[1/2,2]$ after restricting
the neighborhoods. These equations impose the stationary-chord identities
\begin{equation}
 g_{k+1}^\top s_k=0,\qquad y_k^\top s_{k+1}=0
 \label{eq:c6-chords}
\end{equation}
at both parities of the orbit.

\paragraph{One infinite orbit and nonzero steps.}
Choose compact convex normalized-data neighborhoods $K_p$ around $v_0$
inside the transition domains. The limiting outgoing datum is $v_0$ for
both allowed transitions, so the neighborhoods can be chosen compatibly.
Choose a nondecreasing schedule $p_j\to\infty$ with
$p_{j+1}-p_j\in\{0,1\}$. A degree is repeated until the scale majorant
$u_{j+1}=2u_j^{p_j(p_j-1)}$ is small enough for the next transition.
For a finite horizon, fix the terminal datum to $v_0$ and use the compact
convex product of the $K_{p_j}\times[1/2,2]$. Compute the scales from the
input factors $E_j$ and apply each $S_{p_j,p_{j+1}}$ to its scale and next
normalized datum. This defines a continuous self-map of that product;
Brouwer's theorem gives a finite exact orbit.

Fix the same $\varepsilon_1>0$ for all horizons. Define
$b_1=\varepsilon_1$ and
$b_{j+1}=\tfrac12 b_j^{p_j(p_j-1)}$. Every finite orbit satisfies
\[
 0<b_j\le\varepsilon_j\le u_j,\qquad
 \varepsilon_{j+1}=\varepsilon_j^{p_j(p_j-1)}E_j.
\]
The lower bound $b_j$ is independent of the horizon for each fixed $j$.
A coordinatewise diagonal limit therefore preserves positive scales and
all exact transition equations. The resulting points
\[
 x_{2j-1}=P_{p_j}(\varepsilon_j;v_j),\qquad
 x_{2j}=R_{p_j}(\varepsilon_j;v_j)
\]
have norms comparable to $\varepsilon_j^{p_j-1}$ and
$\varepsilon_j^{p_j}$, respectively. Choosing the scale thresholds small
enough makes consecutive norm ratios at most $1/4$. Thus every point
and chord is nonzero, and the single infinite orbit converges to zero.

\newpage
\paragraph{Smooth interpolation with a uniform Hessian bound.}
At an orbit point $z$ of degree $p$, write $\Delta_z=G_p(z)-z$.
The explicit polynomial gradients give
\[
 \|\Delta_{P_p}\|\le C_0^p p^2\|P_p\|^{p+1},\qquad
 \|\Delta_{R_p}\|\le C_0^p p^2\|R_p\|^p,
\]
with $C_0$ independent of the stage. The separated norms allow disjoint
balls $B(z,c\|z\|)$ for a fixed sufficiently small $c>0$.
For a smooth cutoff $\chi$ equal to one near zero and supported in the
unit ball, define
\[
 \psi_z(x)=\chi\!\left(\frac{x-z}{c\|z\|}\right)
                   \Delta_z^\top(x-z),\qquad
 F_2(x)=\tfrac12\|x\|^2+\psi(x),\quad \psi=\sum_z\psi_z.
\]
Each bump satisfies $\nabla\psi_z(z)=\Delta_z$ and, for every fixed
integer $\ell\ge0$,
\begin{equation}
 \|D^\ell\psi_z\|_\infty
 \le C_\ell\|\Delta_z\|\,\|z\|^{1-\ell}.
 \label{eq:c6-bump-bound}
\end{equation}
Since $p_j\to\infty$ and the scales decrease at least by a sixth power
at each transition, these bounds are $o(\|z\|^N)$ for every fixed
$\ell,N$. On each support, $\|x\|$ is comparable to $\|z\|$.
The sum therefore extends smoothly across zero with all derivatives of
$\psi$ equal to zero there. Taking the initial scale and the transition
thresholds sufficiently small also gives
\[
 \sup_x\|\nabla^2\psi(x)\|_{\mathrm{op}}<\tfrac12,\qquad
 \nabla^2F_2(x)\succeq\tfrac12 I_2\quad(x\in\mathbb R^2).
\]
Thus one globally $C^\infty$, strongly convex function realizes all the
prescribed gradients, with $\nabla F_2(0)=0$ and $\nabla^2F_2(0)=I_2$.

\paragraph{Realization by exact Hessian-form BFGS.}
Strong convexity and Eq.~\eqref{eq:c6-chords} give
\[
 s_k^\top y_k=-g_k^\top s_k\ge\tfrac12\|s_k\|^2>0.
\]
Choose $B_1\succ0$ with $B_1s_1=-\lambda_1g_1$ for some
$\lambda_1>0$. Such a symmetric matrix exists because
$s_1^\top(-g_1)>0$: prescribe its first column in a basis beginning with
$s_1/\|s_1\|$, and choose the complementary diagonal entry large enough
for positive definiteness. The BFGS update preserves positive definiteness
and gives $B_{k+1}s_k=y_k$. Hence
\[
 s_k^\top B_{k+1}s_{k+1}=y_k^\top s_{k+1}=0,
 \qquad g_{k+1}^\top s_k=0.
\]
In two dimensions, $B_{k+1}s_{k+1}$ and $g_{k+1}$ lie in the same
one-dimensional orthogonal complement of $s_k$. Positive definiteness
and $g_{k+1}^\top s_{k+1}<0$ fix the sign of their proportionality:
\[
 B_{k+1}s_{k+1}=-\lambda_{k+1}g_{k+1},\qquad\lambda_{k+1}>0.
\]
Thus $s_k=\lambda_kd_k$. The first identity in
Eq.~\eqref{eq:c6-chords} makes $x_{k+1}$ stationary on this search line;
global strong convexity makes it the unique global line minimizer.
Consequently $\alpha_k=\lambda_k$ realizes the prescribed orbit, with
nonzero steps and positive-definite matrices at every finite iteration.

\newpage
\paragraph{Q-order of the full run.}
Let $a_k=-\log\|x_k\|$ and $\ell_j=-\log\varepsilon_j$. The bounded
normalized data and scale factors give
\begin{align*}
 a_{2j-1}&=(p_j-1)\ell_j+O(1),&
 a_{2j}&=p_j\ell_j+O(1),\\
 \ell_{j+1}&=p_j(p_j-1)\ell_j+O(1).
\end{align*}
It follows that
\[
 \frac{a_{2j}}{a_{2j-1}}\longrightarrow1,\qquad
 \frac{a_{2j+1}}{a_{2j}}\longrightarrow+\infty.
\]
For any fixed $p>1$, the first subsequence yields
\[
 \log\frac{e_{2j}}{e_{2j-1}^{p}}
 =\left(p-\frac{a_{2j}}{a_{2j-1}}\right)a_{2j-1}
 \longrightarrow+\infty.
\]
No constant $C$ can therefore make $e_{k+1}\le Ce_k^p$ hold for all
sufficiently late $k$. Together with the universal order-one bound, this
proves $q=1$ for the same infinite run. The log-error increments also tend
to infinity on both parities: $a_{2j}-a_{2j-1}=\ell_j+O(1)$, and
$a_{2j+1}/a_{2j}\to\infty$. Hence $e_{k+1}/e_k\to0$ as well.

\paragraph{Attainment in every fixed dimension.}
For $n>2$, set
\[
 F_n(u,z)=F_2(u)+\tfrac12\|z\|^2,\qquad
 B_1^{(n)}=\operatorname{diag}(B_1^{(2)},I_{n-2}),\qquad
 x_1^{(n)}=(x_1^{(2)},0).
\]
Induction in Eq.~\eqref{eq:c6-update} preserves
$B_k^{(n)}=\operatorname{diag}(B_k^{(2)},I_{n-2})$; all gradients,
directions and steps remain in the first two coordinates. Each line
objective and each error norm equals its planar counterpart. This preserves
exact line search, global smoothness and strong convexity, and Q-order
one. Thus a single attaining run exists in every fixed $n\ge2$, proving
Theorem~\ref{thm:c6}.

\paragraph{Comparison with the classical result.}
\citet{yuan1984least} establishes an infimum of one for exact-line-search
variable-metric methods on convex $C^2$ objectives with a positive-definite
Hessian at the minimizer, using a family of examples with Q-order
$1+1/N$ under the adjacent-iterate power convention in
Eq.~\eqref{eq:c6-qorder}. Its Broyden-family setting includes BFGS, and its two-dimensional
construction is extended to every fixed $n\ge2$. The C6 construction
establishes Q-order exactly one on a single globally $C^\infty$ strongly
convex objective and one infinite exact-BFGS run. Thus the comparison
concerns smooth realization and attainment, with the shared algorithmic
and dimensional scope made explicit in Table~\ref{tab:c6-comparison}.

\begin{table}[htbp]
\centering\small\setlength{\tabcolsep}{4pt}
\caption{C6 compared with the classical least-Q-order result.}
\label{tab:c6-comparison}
\begin{tabularx}{\linewidth}{@{}p{0.16\linewidth}YY@{}}
\toprule
Aspect & \citet{yuan1984least} & C6 construction\\
\midrule
Regularity &
Convex $C^2$ objective; positive-definite Hessian at the minimizer &
Globally $C^\infty$ and strongly convex example in the task's locally
strongly convex class\\
Q-order &
Adjacent-iterate Q-orders $1+1/N$ for arbitrarily large $N$ &
Adjacent-iterate Q-order exactly one under
Eq.~\eqref{eq:c6-qorder}\\
Attainment &
Infimum one obtained from a family of runs &
One infinite run attains the infimum in each fixed dimension\\
Algorithm &
Broyden-family updates, including BFGS, with perfect line searches &
Hessian-form BFGS with global minimization on every search line\\
Dimension &
Every fixed $n\ge2$, by extension of the two-dimensional example &
Every fixed $n\ge2$, preserving BFGS and Q-order under a block lift\\
\bottomrule
\end{tabularx}
\end{table}
\clearpage

\section{Director-led execution and evidence-tracking memory}
\label{app:execution}
\begin{table}[htbp]
\centering\small
\caption{Operational checks and semantic decisions.}
\label{tab:boundaries}
\begin{tabularx}{\linewidth}{lYY}
\toprule
Object & Host checks or records & Model judgment\\
\midrule
Dispatch & Roles, limits, input references & Objective and action selection\\
Revision & Content change, version, review reset & Mathematical meaning of change\\
Reuse & Sources and historical arguments & Applicability and valid inference\\
Review & Same target and current candidate & Scope and argument correctness\\
\bottomrule
\end{tabularx}
\end{table}

\subsection{Execution modes and specialist roles}
The direct proof mode begins with the supplied proposition and develops a
route toward its proof, using local derivations and research as needed to
resolve missing steps. The research program mode begins with a survey of
relevant results and a framing of the original question into mathematical
subproblems. Refined conjecture investigation then develops candidate tasks
and their supporting mechanisms within that framework. Both modes use the
shared director-led loop to select subsequent work from the available
mathematical findings and unresolved obligations.

Research framing records each subproblem's objective, dependencies and
expected result. Candidate refinement and splitting preserve the assigned
mathematical scope, while local findings can reveal further questions for
investigation. An RCI handoff supplies candidate statements, supporting
arguments and precise remaining obligations; an investigation may also end
without an accepted candidate. The director uses these outcomes to select
subsequent research, local-workbench derivations, or whole-proof synthesis,
with completion assessed against the original target.

The director chooses one bounded action using the current target, recent
handoff, verification feedback and available research records. It can
dispatch survey, research, RCI, planner, local workbench, or whole prover.
The role instruction explicitly separates scheduling from proving: the
director should not rederive mathematical arguments while selecting the
next specialist. The complete public reference catalog is available, but
the reading order prioritizes recent evidence instead of requiring the
entire history to be reread on every round.

Table~\ref{tab:roles} summarizes the production roles and call preparation.
The prompt composer prepares a selected call; it is not a mathematical
research specialist dispatched by the director. Individual runs use the
roles needed by their current objectives.
The actual sequence is part of the execution trace and the common execution
budget includes unsuccessful or rejected specialist work.

\begin{table}[htbp]
\centering\small
\caption{Responsibilities and outputs of research roles and call preparation.}
\label{tab:roles}
\begin{tabularx}{\linewidth}{lYY}
\toprule
Component & Main responsibility & Output and authority\\
\midrule
director & Select one next action and its inputs & Scheduling directive; no mathematical verdict\\
prompt composer & Organize a selected task and its mathematical inputs & Role-specific prompt; no scheduling decision or mathematical verdict\\
survey & Acquire and inspect permitted sources & Source records and applicability gaps\\
research framing & Organize the original request & Bounded work items and dependencies\\
RCI & Develop candidates and support & Candidate tasks, mechanisms, uncovered regions and refutations\\
planner & Develop or substantially revise a route & Dependencies and unresolved bridges\\
research & Investigate a bounded mathematical objective & Search findings for further work\\
local workbench & Address a precise proof gap & Scoped note and first remaining obstacle\\
whole prover & Write the complete argument for the target & Candidate awaiting verification\\
scope auditor / math reviewer & Review scope and mathematics & Bound internal verification evidence\\
\bottomrule
\end{tabularx}
\end{table}

\subsection{Prompt composer: interface and selective invocation}
\label{app:composer}
Let $\mathcal S$ be the fixed method catalog and $\mathcal S(a)$ its
role-eligible subset. A directive
$d_t=(a_t,g_t,R_t,S_t)=\operatorname{director}(T_0,M_t,\mathcal S)$ specifies
role $a_t$, objective $g_t$, records $R_t\subseteq M_t$ and methods
$S_t\subseteq\mathcal S(a_t)$. Context preparation follows
\begin{equation}
p_t=\begin{cases}
\operatorname{composer}(T_0,d_t),&\chi(d_t)=1,\\
\operatorname{assemble}(T_0,d_t),&\chi(d_t)=0.
\end{cases}
\label{eq:composer}
\end{equation}
The fixed policy $\chi$ selects the calls described below. Both paths supply
the specialist with the same input files and methods,
$z_t=\operatorname{specialist}_{a_t}(p_t,R_t,S_t)$.

The prompt composer receives the selected objective, role, supplied material
references and selected skills $S_t$ drawn from the role-eligible catalog
$\mathcal{S}(a_t)$. These inputs are chosen by the director;
startup calls use the corresponding initialization assignment. The prompt composer
organizes their mathematical content for execution as described in
Section~\ref{sec:seg-memory}. Invocation follows fixed role and stage rules.
The survey role owns its composition
call; initial framing, new or substantially rewritten plans and
tasks requiring a complete argument also use the prompt composer. Research, local workbench and
fixed RCI phases consume their role guidance and input references directly.
Representation-only corrections preserve the existing task instead of
requiring a new semantic compilation of it.

The prompt composer is instructed to reconcile the literal target, legal
assumptions, source evidence and requested deliverable. Its output
clarifies the target and its assumptions, relates available sources to the
task and identifies unresolved steps, falsification checks, uses of the selected skills,
and required outputs. The specialist uses this context to choose and develop
an argument.

The method catalog is fixed during evaluation. Skills prepared for each run
provide reusable mathematical workflows for
the current role. They guide decomposition, source transfer,
counterexample construction, boundary exploration, object or algorithm
construction and quantitative refinement. Their availability does not
turn their example arguments into premises for the current target.
Applicability still requires matching the current domain, quantifiers,
conditions and requested guarantee.

\paragraph{Core instruction and input format.}
The prompt composer receives the following mathematical instructions
(excerpt):
\begin{quote}\small
You organize a work prompt for one mathematical role. You do not solve the
problem, choose a mathematical route, or declare a result.

Reconcile the literal target, legal assumptions and requested deliverable
with the source records. For each source, state its exact claim, hypotheses,
locator, authority and the target clause it may inform. Mark missing
applicability bridges and failed routes as open search work.

Organize the final prompt as: target and scope $\to$ deliverable $\to$
source-to-target mapping $\to$ route and proof frontier $\to$ independent
attack or falsification check when useful $\to$ Cards/Skills $\to$ output
protocol.
\end{quote}
The request supplies a stage and role, an objective brief, references to the
target and research materials, the selected method guides and the required
output destination. These are references to existing files rather than a
replacement summary of their contents. The returned text is the complete
call-specific prompt; its ordered sections express the mathematical
relationships above. Call provenance is recorded separately. Both this
prompt and the original supplied files reach the specialist. Fixed RCI
stages instead use their own role instructions and stage-specific methods.

\subsection{Recorded composite-optimization investigation}
\label{app:l0l1-running-example}
The example in Section~\ref{sec:conjecture-example} uses the completed C1
investigation. Its literal request
separates strong and weak subgradient regimes, upper and lower complexity
bounds, regional legality, zero parameters and oracle counts. It permits a
proof, refutation, or precise additional conditions for the requested rate.

\paragraph{Refining the mathematical target.}
Seed review replaces an instance-specific method with the requirement of
one uniform algorithm and restricts this investigation to the positive
strong/global regime. Candidate review then resolves an ambiguity in the
complexity claim. For fixed parameters $\theta=(L_0,L_1,\ell,R)$, the bound
has the form $N\le C R\sqrt{L_{\mathrm{eff}}/\varepsilon}
+r_\theta(\varepsilon)$, where $C$ is universal and
$\sqrt{\varepsilon}\,r_\theta(\varepsilon)\to0$ as $\varepsilon\to0$.
The same algorithm and bound must work uniformly over legal instances and
dimensions. Making these quantifiers explicit also fixes the corresponding
refutation target: a failure of one proposed algorithm or intermediate
estimate is insufficient to refute the uniform existence claim.

\paragraph{Reassessing a useful local result.}
Before this candidate revision, mechanism review had retained the radial
curvature certificate $\bar L_\rho=e^{L_1\rho}L_{\mathrm{eff}}$ as partial
support. Updating the candidate clears that mechanism's current support
status and the associated coverage assessment. The subsequent investigation
proposes the certificate for the revised candidate and obtains fresh
mechanism and coverage reviews. The derivation remains valid, but applying
it to the requested rate still requires all relevant query points, proximal
outputs and model segments to lie within the certified region, with
$L_1\rho=O(1)$. The renewed coverage assessment retains partial support and
identifies trajectory control as the next proof objective. Evidence-tracking
memory preserves the conditional result and the open connection to the
candidate; context preparation passes both to subsequent proof development.

\begin{figure}[htbp]
  \centering
  \includegraphics[width=\linewidth]{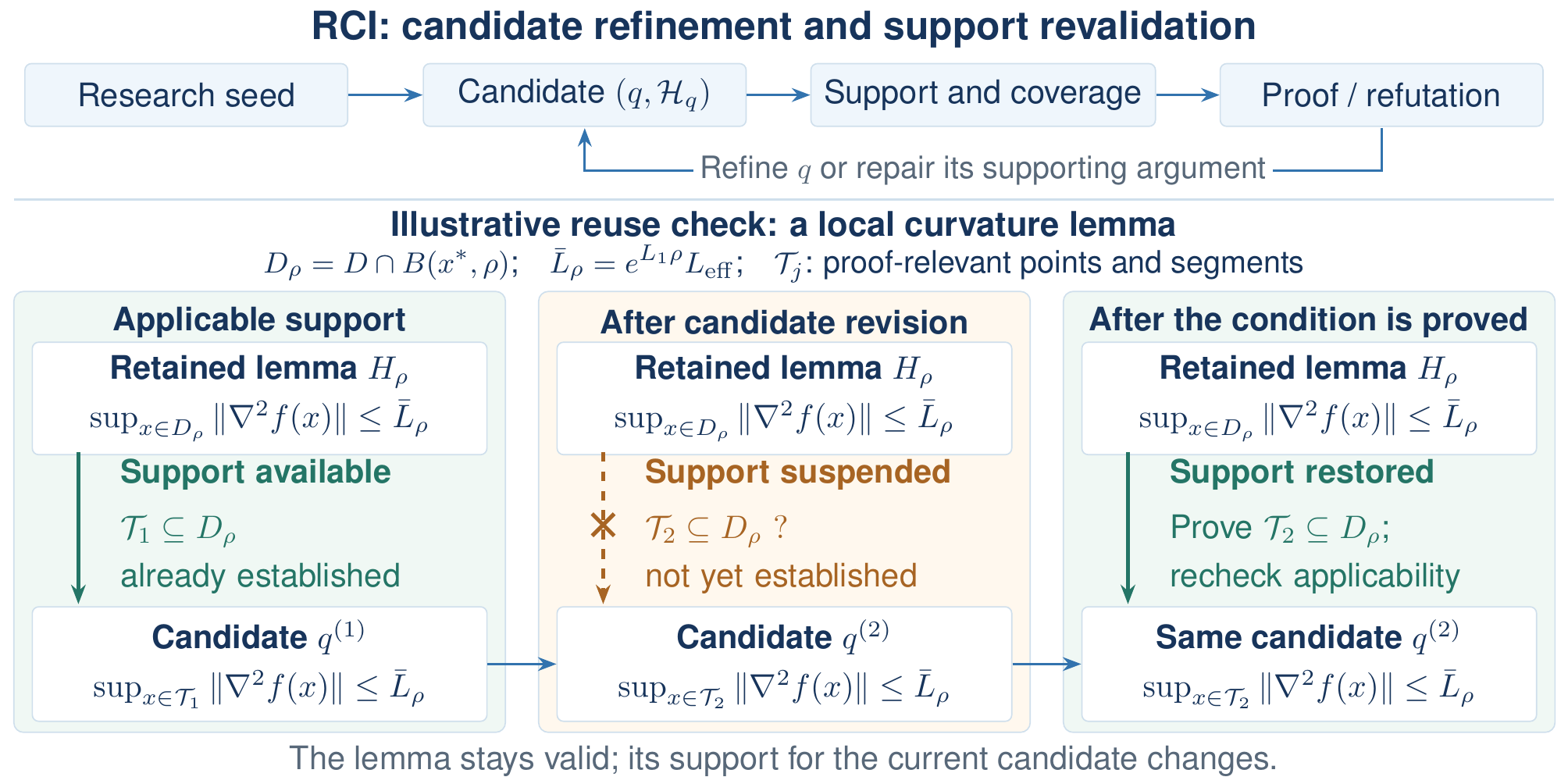}
  \caption{Illustrative support revalidation in RCI: the lemma persists,
  while a revised candidate requires a new containment proof.}
  \label{fig:revision-support}
\end{figure}

Figure~\ref{fig:revision-support} gives a schematic example of this distinction.
The retained lemma $H_\rho$ bounds curvature on
$D_\rho=D\cap B(x^*,\rho)$. Its application to a candidate $q^{(1)}$
requires the relevant points and segments $\mathcal T_1$ to lie in $D_\rho$.
If the revised candidate $q^{(2)}$ concerns a different set $\mathcal T_2$,
the earlier containment argument does not establish
$\mathcal T_2\subseteq D_\rho$. The lemma remains valid, but its support
for $q^{(2)}$ is suspended until this new condition is proved and its
applicability checked. The reopened support relation thus identifies a
specific mathematical objective for further investigation.

\paragraph{Observed revision process.}
The records contain one seed revision, three candidate versions and two
separately reviewed mechanism records. One candidate change refines the
complexity quantifiers; the other changes the transcription of the context
and assumptions while leaving the statement and its positive and refutation
targets unchanged. The substantive revision clears one existing mechanism's
support status. After the revised candidate is formulated, one mechanism
review, one coverage assessment and one final candidate review take
26.7 minutes in the recorded execution. These observed review calls and elapsed time characterize the recorded
reassessment sequence.

\paragraph{Developing the remaining proof obligations.}
RCI returns partial findings with the trajectory condition unresolved, and
the research program turns that condition into further mathematical objectives.
The conditional-upper deliverable derives the proximal three-point inequality,
telescoping Lyapunov bound, rate and exact gradient/proximal counts under
executed quadratic models and regional legality. Subsequent research in the strong/global subgradient regime establishes the bound
$\bar L=e^{1+L_1R}L_{\mathrm{eff}}$ along all relevant points and segments of the certified trajectory. The final
candidate also reports a weak-regime finite-budget oracle obstruction and a
lower bound obtained from a hidden-rotation quadratic and affine composite
lift. The accelerated upper bound applies under bounded $L_1R$ or the stated
trajectory certificate, while the lower-bound construction concerns
deterministic algorithms.
The final report records \texttt{PASS}/\texttt{VERIFIED}; the scope audit and
mathematical review bind the same final candidate and accept its stated
regime-by-regime answer, including these qualifications. The trace links
candidate refinement to support reassessment and then to new proof work:
the retained local estimate supplies a conditional premise, while its
unresolved applicability determines what must be established next.

\subsection{Memory, Failed Routes and Source Applicability}
Evidence-tracking memory stores explicit artifacts: source records, plans, candidate
proofs, local notes, attempted routes and verification feedback. Memory is
updated by persisting versioned research artifacts and review feedback with
provenance (Section~\ref{sec:loop}). With specialist output $z_t$ and review
feedback $\mathcal V_t$, the record update is
$M_{t+1}=M_t\oplus\{z_t\}\oplus\mathcal V_t$.
A record can be useful because it rules out a
route or identifies an unproved bridge, even when it contains no new theorem.
The director uses that information to choose the next objective, while
context construction makes the relevant mathematical details available
to the selected role.

A failed route should identify the exact step, the assumptions active at
that step and the reason the proposed inference is unavailable. A failure
caused by an unverified source hypothesis differs from a counterexample
to the original conclusion. The director uses these provisional diagnoses to
schedule correction of source theorem applications, mechanism exploration, or target-level
refutation.

Source reuse requires matching the current domain, quantifiers, parameter
regime and hypotheses. Section~\ref{sec:research-outcomes} formalizes these checks;
an analogy or numerical pattern alone does not establish applicability.

\subsection{Bounded execution}
Scheduling rounds, research calls and complete-proof attempts have separate
budgets. The director selects one specialist at a time, and components may
have bounded internal phases. The execution record retains the selected
objectives, mathematical outputs and verification results. The final output
contains either an accepted candidate or the established partial results
and remaining obligations.

\section{RCI Phases and Research-Task Revision}
\label{app:smi}

\paragraph{Reuse after revision.}
Revision changes the applicability of established results, not their truth.
Given a proved statement $\forall u\in\mathcal X,\ \Gamma(u)\Rightarrow\varphi(u)$,
sufficient bridges to revised assumptions $\Gamma'$ and conclusion $\varphi'$ are
\begin{equation}
\begin{aligned}
 &\forall u\in\mathcal X,\quad \Gamma'(u)\Rightarrow\Gamma(u),\\
 &\forall u\in\mathcal X,\quad
 \Gamma'(u)\land\varphi(u)\Rightarrow\varphi'(u).
\end{aligned}
\label{eq:candidate-transfer}
\end{equation}
Missing bridges block this reuse, not other proofs. Unproved old support
still requires validity and applicability checks.
An old counterexample $u^\star$ refutes
the revised candidate only if $\Gamma'(u^\star)\land\neg\varphi'(u^\star)$.

\subsection{Representative Task Forms}
\label{app:task-forms}
For objects $u\in\mathcal X_q$, the universal candidate and its negation are
\begin{align}
 &\forall u\in\mathcal X_q,\quad\Gamma_q(u)\Rightarrow\varphi_q(u),
 \label{eq:positive-task}\\
 &\exists u\in\mathcal X_q,\quad\Gamma_q(u)\land\neg\varphi_q(u).
 \label{eq:refutation-task}
\end{align}
A construction in admissible class $\mathcal A_q$ has representative form
\begin{equation}
 \exists a\in\mathcal A_q\;\forall u\in\mathcal X_q,\quad
 \Gamma_q(u)\Rightarrow\varphi_q(u;a).
 \label{eq:construction-task}
\end{equation}
Its negation requires failure for every admissible $a$, not one selected
construction. Deliverables specify whether an explicit witness, executable
algorithm, or complexity guarantee is needed. Additional quantifier alternations
retain their original dependencies.

\subsection{Seed, Frontier and Mechanism Review}
For a selected ready work item, seed scout proposes a mathematical seed
within the original request. Separate raw-seed reviews examine correctness
risks and research value. Their roles differ: a statement can be meaningful
but false, or mathematically plausible but merely repeat an available
source. Bounded seed repair addresses the explicit criticism before
frontier exploration. The current phase graph uses fixed role prompts
with stage-specific method guidance, rather than a prompt composer call at each
internal step.

Frontier explorer develops a precise candidate, its potential falsifiers,
and supporting mechanisms. A targeted refutation probe can examine fragile
parts of the proposed boundary. Mechanism review asks whether the proposed
support is applicable, whether it relies on the conclusion being sought,
and what mathematical content remains missing. The support and coverage relations in the main text express these
review obligations.

Coverage review distinguishes a substantive partial result from apparent
complete coverage. A useful mechanism may close one necessary step without
establishing the target. The phase can request a bounded additional support
move, revise the frontier, or direct an apparent resolution toward further
checking. Candidate splitting and repair are constrained by the phase
budget, so the proposal graph does not expand indefinitely.

\subsection{Challenge, Probes and Deduplication}
Final challenger reassesses the candidate after mechanism and coverage
review. Proof and refutation probes then attempt to expose an overlooked
solution, contradiction, or ill-posed boundary using mechanism-free views.
A separate frontier probe can use a source-free view. Caller-supplied predecessor references
remain available to these probes.

Open candidates pass semantic deduplication before being returned as
examined candidate-task contexts suitable for further work. A candidate apparently resolved by a proof or
counterexample is instead retained as a task for checking a proposed resolution. These two
paths differ in what should be checked next, not in whether the original
request has already been solved. Both return to the same director for
research, planning, local reasoning, or construction of a complete argument.

RCI can return one or two examined candidate-task contexts under the current bounded design,
an outcome with no task accepted by RCI, or an interface failure. These contexts preserve
the candidate statement, legal assumptions, outstanding obligations and
relevant mechanism references. RCI task acceptance marks a context as ready for subsequent work.
Internal proof acceptance follows the separate scope-auditor/Mathematical
Reviewer procedure (the latter uses the implementation name kernel).

\subsection{Revision Dependencies and Diagnostic Outcomes}

Candidate and mechanism records carry identifiers, content references and
versions. A candidate commit clears promoted support, coverage, challenge,
and frontier-probe references; a mechanism repair returns the changed support
to its critic. Earlier arguments remain available under their original
assumptions. Reuse follows (\ref{eq:candidate-transfer}); proposed
support still requires proof and old counterexamples must satisfy the revised
assumptions. The host resets candidate-linked review records after revision,
while mathematical review determines the applicability of retained results
across work items.

\begin{center}\small
\begin{tabularx}{\linewidth}{@{}lX@{}}
\toprule
Revision & Required reassessment\\
\midrule
Mechanism only & Keep the candidate; recheck changed support and affected coverage, challenge and probes.\\
Candidate statement & Recheck uses of support and counterexamples under the revised assumptions, conclusions, or boundaries.\\
Representation only & Preserve mathematical content; a semantic change requires substantive reassessment.\\
\bottomrule
\end{tabularx}
\end{center}

Representation repair preserves the mathematical task. Changes to
assumptions or conclusions require substantive reassessment. An investigation
that returns no candidate leaves its mathematical objective unresolved.

\section{Verification and Evaluation Details}
\label{app:verification}
\subsection{From Local Reasoning to an Accepted Candidate}
\label{app:local-transfer}
The local workbench addresses a precise mathematical gap using the original
target, current proof and relevant verification feedback. Its note states
what was derived, under which assumptions it applies and which bridge
remains unresolved. The note is a local research artifact: it records the derivation, its
assumptions and the remaining bridge. The planner similarly proposes
dependencies and routes, while the whole prover turns the selected materials
into a self-contained candidate for the original deliverable.
For a target $\forall\lambda\in D,\Phi(\lambda)$, local results on $D_i$
require $D\subseteq\bigcup_iD_i$ together with each result's hypotheses.

For local finding $i$, let $\Gamma_i(v)$ and $\varphi_i(v)$ denote its
assumptions and conclusion for $v\in\mathcal{V}_i$. At its point of use,
$u\in\mathcal{U}_i$ describes the current objects and $\Delta_i(u)$ the
available assumptions. A substitution $\sigma_i:\mathcal{U}_i\to\mathcal{V}_i$
connects the domains. Reuse requires
\begin{equation}
\begin{aligned}
&\forall u\in\mathcal{U}_i,\quad
\Delta_i(u)\Rightarrow\Gamma_i(\sigma_i(u)),\\
&\forall v\in\mathcal{V}_i,\quad
\Gamma_i(v)\Rightarrow\varphi_i(v).
\end{aligned}
\label{eq:local-transfer}
\end{equation}
The first establishes applicability and the second the local result;
together they yield $\Delta_i(u)\Rightarrow\varphi_i(\sigma_i(u))$.

The candidate file, not a conversational success message, is passed to
the verification boundary. The proof contract auditor examines target
coverage, allowed assumptions and the relationship between the requested
and demonstrated conclusions. The mathematical reviewer (math verifier kernel
in the implementation) examines the
mathematical argument. Both assessments must refer to the same current
candidate and literal target. A revised proof cannot inherit acceptance
from an earlier version merely because its topic or filename is similar.
The auditor receives the complete original problem, extracted target record,
orientation brief, current argument and source material. Its instruction
requires reading the literal problem independently and treats the brief as
non-authoritative. Thus orientation omissions can be reconsidered during
review, although detecting them remains a semantic judgment.

Internal proof acceptance requires the auditor's acceptance, the mathematical reviewer's
acceptance and explicit eligibility and no unresolved blockers under
the bound result. These are natural-language judgments. Section~\ref{sec:research-outcomes} distinguishes local research outcomes
from internal acceptance. Benchmark review and expert assessment are conducted separately from the system’s internal acceptance procedure.

\subsection{Base-Model Prompts}
\label{app:baseline-prompts}
The following templates specify the unified Frozen protocol. Exact prompts
for reported runs remain recorded with their execution artifacts. Both
baselines access literature through the same Frozen \texttt{search} tool;
no \texttt{open} or live retrieval tool is exposed.

\paragraph{direct prompting.}
This template is supplied once, followed by the unchanged task statement.
Search results enter the model's continuation under the common access rules.

\begin{quote}
\small
Give a proof of the following statement.\\
Use the available Frozen search tool as needed.

[unchanged benchmark statement]
\end{quote}

Both baselines use \texttt{gpt-5.6-sol} with \texttt{xhigh} reasoning, the same
statement and the same paper-level cutoff policy. Each task begins with a
clean conversation.
The self-check prompting template below adds instructions for drafting,
checking and revision while preserving the same search permissions.
The exact prompt, statement, model configuration and retrieval record
identify each attempt.

\paragraph{self-check prompting.}
\label{app:self-revision}
This baseline receives the fixed short prompt below, followed directly by
the unchanged task. Each attempt consists of one user-initiated invocation,
including MCP calls and the model continuations needed to consume their
results. The model controls its drafting, checking and revision within this
invocation; the host does not schedule separate stages for these activities.
The host supplies tool results only: no separate verifier feedback, human
correction or additional user instruction is returned during generation.

\begin{quote}
\small
Solve the mathematical task below. Draft a candidate proof, check it and
revise it as needed before presenting your final answer. Check the conclusion,
assumptions, quantifiers, main inferences and applicability of cited theorems.
If you find a gap or counterexample, repair the argument or try another approach
and check it again. Use your existing knowledge and the available search tool.
Present your final proof.

[unchanged benchmark statement]
\end{quote}

One task execution ending in a final answer counts as one attempt, including
its Frozen search calls and tool-result continuations. The model chooses how
much internal revision to perform. Both baselines have a 32-turn ceiling for
model continuations; the first final answer ends the attempt. The only exposed
tool is \texttt{search} and the gateway records each query and returned
historical version. Truncated responses remain recorded outcomes and receive
no manual continuation. Independent benchmark review follows generation.

\subsection{Experimental Protocol and Reproducibility}
\label{app:exp-protocol}
\label{app:ablations}

\paragraph{Execution settings.}
Each run records its model, reasoning setting, implementation revision, role
instructions, visible inputs, tools, access policy and execution limits.
The known-proof comparison appears in
Table~\ref{tab:exp-direct}; conjecture research includes the six-case
comparison in Table~\ref{tab:exp-conjecture} and the five C1 workflow ablations
in Table~\ref{tab:exp-components}. Each C1
configuration retains one evaluated submission. The mechanistic rationale and
its relation to the recorded behavior are given in
Appendix~\ref{app:mechanistic-rationale}.

\paragraph{Resource accounting and planned allocations.}
All methods use the same accounting unit: cumulative input plus output
tokens across agent-side model calls, including initialization, prompt
composition, specialist reasoning, internal verification, critique, revision
and RCI. Repeated and cached input count on every call; reasoning tokens
are included in output. Consumed tokens from retries and failed calls are
included when available. Independent post-hoc evaluation is accounted for
separately. The five answers and four critiques of iterative solving belong
to one task attempt.

Table~\ref{tab:token-budgets} gives the initial planned task allocations.
Table~\ref{tab:token-consumption} reports generation-token consumption for
C1--C6, the C1 ablations and the known-proof comparison. Both tables report
millions of tokens (M), rounded to two decimal places. Planned allocations
and consumption across the full generation workflow are reported separately.
\begin{table}[htbp]
\centering\small
\caption{Originally specified generation allocations per task.
These planned amounts are distinct from the consumption reported below.}
\label{tab:token-budgets}
\begin{tabularx}{\linewidth}{@{}lYr@{}}
\toprule
Task or class & Proof scope & Planned tokens (M)\\
\midrule
Known-proof S1 & Local derivation, certificate, or short key lemma & 3.00\\
Known-proof S2 & Single theorem with several connected proof steps & 5.00\\
Known-proof S3 & Sharp bound, long dependency chain, or new construction & 8.00\\
\midrule
C1 & Composite optimization: regime analysis and upper/lower bounds & 20.00\\
C2 & Finite-chain and statistical lower bounds; sample normalization & 8.00\\
C3 & Tree aggregation, accelerated potential, and propagation bound & 6.00\\
C4 & Time-varying consensus, tracking, and coupled descent bounds & 12.00\\
C5 & Stochastic and directed-network hard instances; Perron analysis & 10.00\\
C6 & Infinite BFGS orbit, smooth interpolation, and dimension extension & 12.00\\
\bottomrule
\end{tabularx}
\end{table}

\begin{table}[!t]
\centering\small
\caption{Generation-token consumption for C1--C6, C1 ablations and known-proof tasks.}
\label{tab:token-consumption}
\begin{tabularx}{\linewidth}{@{}Yr@{}}
\toprule
Method or case & Tokens (M)\\
\midrule
\multicolumn{2}{@{}l}{\textit{C1: full system, ablations and baselines}}\\
\agent\ & 18.46\\
A1: Single-pass research--proof & 7.96\\
A2: Fixed-prompt context organization & 14.69\\
A3: Predetermined research sequence & 13.64\\
A4: Proof critique and revision & 20.60\\
A5: Without the revision-triggered revalidation module & 14.29\\
C1 Direct & 1.12\\
C1 Iterative & 12.10\\
\midrule
\multicolumn{2}{@{}l}{\textit{C2--C6: full system and baselines}}\\
C2 Direct & 0.06\\
C2 Iterative & 7.12\\
C2 \agent\ & 7.92\\
C3 Direct & 0.03\\
C3 Iterative & 5.10\\
C3 \agent\ & 5.50\\
C4 Direct & 0.04\\
C4 Iterative & 10.21\\
C4 \agent\ & 11.50\\
C5 Direct & 0.02\\
C5 Iterative & 8.97\\
C5 \agent\ & 9.70\\
C6 Direct & 0.07\\
C6 Iterative & 11.11\\
C6 \agent\ & 11.60\\
\midrule
\multicolumn{2}{@{}l}{\textit{Known-proof: per task}}\\
Direct & 0.18\\
Self-check & 1.28\\
QED & 4.59\\
Rethlas & 4.44\\
Iteris & 5.59\\
\agent\ & 5.32\\
\bottomrule
\end{tabularx}
\end{table}
\FloatBarrier

\paragraph{Native stopping rules and submitted outcomes.}
\label{app:stopping-rules}
Realized consumption depends on the workflow, context lengths, number of
model calls and stopping decisions. QED, Rethlas, Iteris and \agent\
retain their native research, verification and submission procedures.
The agent-baseline comparison evaluates the result delivered under each
system's submission rules. For known-proof prompting, the first final
answer ends the attempt, including self-checking within that invocation.
\agent\ continues research and proof development until its termination
condition or an execution limit is reached.
Independent mathematical review assesses the submitted outputs against
the original task.

\paragraph{Single-invocation reasoning and sustained research.}
Direct and self-check prompting perform their mathematical work within one
user-initiated invocation, including search-tool continuations. Their
consumption depends on the reasoning and revision the model performs before
returning its final answer. Increasing a nominal task allocation does not
itself schedule further mathematical work after that answer. Continuing
research requires a workflow that retains the current proof state,
identifies unresolved objectives, dispatches further reasoning and feeds
review findings into subsequent revisions. This motivates agent design:
\agent\ organizes an extended task budget into successive research,
verification and proof-development calls. Comparisons with prompting
baselines therefore assess complete workflows, including their allocation
of computation and stopping decisions.

\paragraph{C1 task and execution settings.}
All C1 configurations use the same statement and permitted public source access.
Mathematical roles use \texttt{gpt-5.6-sol} with \texttt{xhigh} reasoning,
and the director uses its \texttt{low} setting. Generation excludes the
full-system proof, other submissions and evaluation materials.
Resource accounting follows the common definition above.
Each variant retains the full-system procedures except for its stated changes. Each selected submission receives
one anonymous \texttt{gpt-6-astra}/\texttt{xhigh} review and separate human
assessment under the original C1 0--7 rubric in Appendix~\ref{app:open}.

\paragraph{Single-pass research--proof.}
\label{app:a1-record}
\emph{Intervention.} Each preparation and specialist role, including each
RCI phase, executes at most once. Candidate and mechanism revisions are
disabled, and initial research framing omits prompt composition.
\emph{Retained capabilities.} The planner and whole prover retain prompt
composition; method guidance and permitted sources remain available.
\emph{Execution and submission.} The director selects among unused roles,
and RCI returns the findings and obligations from this single pass.
One complete proof is constructed and independently verified before
generation ends.

\paragraph{Fixed-prompt workflow.}
\label{app:a2-record}
\emph{Intervention.} Fixed role prompts replace task-specific prompt
composition, changing only how the mathematical context is organized.
\emph{Retained capabilities.} The task, current objective, available evidence,
method guidance, resources, roles and tool permissions are held fixed.
Adaptive research selection, RCI, proof iteration and final verification
remain available under the same model settings and execution limits.
\emph{Execution and submission.} Fixed role prompts present the shared
information in a static structure. Task-specific composition organizes it
around the current objective, distinguishing applicable checked premises,
exploratory material and unresolved proof obligations, and making
source-to-target applicability explicit. The comparison isolates the role
of task-specific context organization.

\paragraph{Predetermined research sequence.}
\label{app:a3-record}
\emph{Intervention.} A fixed sequence replaces adaptive role selection.
\emph{Retained capabilities.} Prompt composition, method guidance,
evidence-tracking memory and within-role investigation remain available.
\emph{Execution and submission.} After initialization, the agent cycles
through survey, RCI, research, planner, whole prover and verification.
The initialization survey is omitted from the first cycle; research handles
local derivations. Roles receive the current objective and latest feedback,
and the planner revises the strategy at its scheduled position. The run
submits the accepted proof or the last candidate at budget exhaustion.

\paragraph{Proof critique and revision.}
\label{app:a4-record}
\emph{Intervention.} A solver--critic loop replaces RCI's separate candidate
and mechanism refinement and targeted proof and refutation probes.
\emph{Retained capabilities.} Prompt composition outside this module,
evidence-tracking memory, other research roles, whole-proof iteration and
final verification remain available.
\emph{Execution and submission.} The solver drafts an argument, the critic
examines it, and the solver revises it using the critique. The resulting
note returns its conclusions, conditions and open gaps to the research
program for subsequent proof development. This replacement changes the
module's internal reasoning allocation.

\paragraph{Without the revision-triggered revalidation module.}
\label{app:a5-design}
\emph{Purpose.} Dependency-guided revalidation allocates renewed
mathematical review to support relations affected by a candidate revision.
The module uses the changed assumptions, quantifiers and target conclusions
to select applicability checks, while retaining unaffected results under
their established conditions. A5 evaluates the contribution of this
review-allocation module to the quality of the final submission.
\emph{Intervention.} The module is removed as a single functional unit.
\emph{Evaluation.} The final selected candidate is submitted for evaluation
under the original C1 rubric; Table~\ref{tab:components} reports its LLM
and human scores alongside those of the full system.
\label{app:revalidation-execution}

\paragraph{Open-case baseline inputs and final-output rules.}
Both baselines receive the same original task and public background materials
as \agent, with access to the same permitted information sources.
The following role-neutral instructions specify their workflows.
Each baseline starts independently and receives neither the existing agent
proof nor the other baseline's output. These prompts are specific to the open case; the Known-Proof
prompts in Appendix~\ref{app:baseline-prompts} remain unchanged.

\textbf{direct solving (one attempt).}
\begin{quote}\small
Solve the supplied mathematical research problem. State the exact conclusion
and all assumptions. If the request leaves conditions or definitions open,
make any proposed completion explicit and explain its relation to the original
question. Supply a self-contained proof with justified source applications,
and identify every unresolved requirement. Use only the permitted public
materials; record citation locations and access dates. Return one final answer.
\end{quote}

\emph{generic iterative solver main prompt.}
\begin{quote}\small
Solve the supplied mathematical research problem as a standalone answer. State
the exact conclusion and all assumptions. If the request leaves conditions or
definitions open, make any proposed completion explicit and explain its relation
to the original question. Supply a self-contained proof with justified source
applications and identify every unresolved requirement. Use only the permitted
public materials; record citation locations and access dates. Return one
standalone answer for later critique and revision.
\end{quote}

\textbf{generic iterative solver (independent critique and revision).}
D and M use the same base model, reasoning setting and public materials.
M independently produces draft $A_1$ using the generic iterative
solver main prompt above. For each fixed cycle $j=1,\ldots,4$, a separate critic
call receives the original task card, current standalone answer $A_j$ and its cited public
materials. It receives no solver-private reasoning or other system's output.
Its review $C_j$ is then passed to the solver with the task card, $A_j$ and
M's own preceding revision/source notes to produce $A_{j+1}$.
Thus M comprises one initial solver call and four critic--solver pairs; it has
no early acceptance stop and only $A_5$ is evaluated.

\emph{Critic master prompt.}
\begin{quote}\small
Independently audit the current answer against the supplied research task.
Check: (1) the exact conclusion, quantifiers and fixed requirements;
(2) all added assumptions, whether they complete an open condition or narrow a
fixed target and whether the result remains nondegenerate; (3) each substantive
inference, boundary case and parameter dependency; (4) cited results and their
applicability; and (5) every missing step between the proved claims and the
claimed completion. For each issue give its location, mathematical reason,
and required correction. Distinguish a demonstrated error from an unresolved
question. Return a mathematical critique, not a replacement proof or outcome score.
\end{quote}

\emph{Solver revision master prompt.}
\begin{quote}\small
Revise the current answer using the supplied mathematical critique and original
task card. Check each criticism rather than treating it as an established fact.
Repair justified issues and explain any rejected criticism using mathematical
evidence. Return a new standalone final answer stating its exact conclusion,
all assumptions, complete proof, citations and unresolved requirements.
Identify which added conditions complete open parts of the question. Earlier
answers and reviews cannot supply an omitted proof step in this answer.
\end{quote}

The task, current answer, critique and public sources occupy separate input
slots. The iterative solver maintains its own reasoning history and starts
independently of direct solving. Only its fifth answer is evaluated.

\paragraph{Evaluation records.}
Appendix~\ref{app:rubric} defines the Known-Proof review, missing-outcome
handling and deterministic score replay. Appendix~\ref{app:open} contains the
open-case evaluator prompt, model configurations, output schema and
human-review procedure. Each selected final submission is linked to its task,
configuration and accepted review; intermediate calls remain execution
records rather than separate experimental outcomes.

\subsection{Recorded Examples of Revalidation and Context Composition}
\label{app:mechanistic-rationale}
The completed C1 investigation records how revalidation and context
composition connect a retained mathematical result to further proof work
(Appendix~\ref{app:l0l1-running-example}). The ablation configurations appear in Appendix~\ref{app:exp-protocol}.

\paragraph{Workflow outcomes and A5 interpretation.}
\label{app:ablation-interpretation}
The full-system proof combines conditional accelerated upper bounds, a
weak-regime obstruction and deterministic lower bounds, retaining the
conditions attached to each conclusion. Single-pass research--proof leaves
several objectives unresolved, while the fixed-sequence and
critique--revision variants achieve broader coverage with gaps in local
justification or sufficient conditions.

The A5 comparison evaluates dependency-guided revalidation as an internal
review-allocation module (Appendix~\ref{app:a5-design}). On C1, the full
system receives LLM and human scores of 7 and 6, while the configuration
without this module receives 3 and 2. The accompanying full-system trace
shows how dependencies guide the choice of mathematical checks: a change
in the candidate's quantifiers reopens the use of a retained curvature
bound, and review identifies the trajectory conditions required to apply
it. Those conditions become explicit objectives for subsequent proof work.
The score comparison reports the workflow outcome; the trace identifies
the mathematical check selected by the module.

\paragraph{Revalidation opens an applicability obligation.}
After the candidate's complexity quantifiers change, the radial-curvature
certificate $\bar L_\rho=e^{L_1\rho}L_{\mathrm{eff}}$ remains valid.
The revalidation module reassesses its applicability to the revised target
and identifies the missing bridge:
query points, proximal outputs and model segments must remain in the
certified region, with $L_1\rho=O(1)$. Trajectory containment becomes an
explicit proof objective. Subsequent proof development derives conditional
upper bounds and a continuation argument under stated conditions. This
recorded sequence shows how reassessment converts the applicability of an
existing result into a concrete task for further investigation.

\paragraph{Context composition carries the premise and its open bridge.}
For the same certificate, the composed context retains the proved local
bound and its region of validity, while presenting trajectory containment
as an unresolved obligation. The next role receives both the available
premise and the condition required to apply it to the current target.
Subsequent proof work derives a conditional convergence bound and supplies
a trajectory certificate; the remaining condition stays attached to the
final conclusion. The ablation retains the same task inputs, source-access policy, method catalog, model settings and execution limits, while replacing task-specific prompt composition with fixed role prompts.
The example identifies the mathematical
information organized by the composer: the certificate's scope and its
source-to-target applicability requirement.

\end{document}

%% file: authors.tex
\author{%
  {\bfseries
    Shu Miao\textsuperscript{1,*}\quad
    Feiming Wang\textsuperscript{2,*}\quad
    Wentao Long\textsuperscript{3}\quad
    Chenyi Li\textsuperscript{1}\quad
    Kun Yuan\textsuperscript{1}\quad
    Zaiwen Wen\textsuperscript{1,\textdagger}\endgraf
  }\vspace{2pt}
  {\small
    \textsuperscript{1}Peking University\quad
    \textsuperscript{2}Nankai University\quad
    \textsuperscript{3}Fudan University\endgraf
  }%
}